\pdfoutput=1

\documentclass[11pt]{article}

\usepackage[]{acl}

\usepackage{times}
\usepackage{latexsym}
\usepackage{amsmath} 
\usepackage{booktabs}
\usepackage{multirow}
\usepackage{subcaption}
\usepackage{colortbl}
\usepackage{tablefootnote}
\usepackage{svg} 

\usepackage[T1]{fontenc}

\usepackage[utf8]{inputenc}
\usepackage{subcaption}

\usepackage{microtype}
\usepackage{hyperref}
\usepackage{inconsolata}

\usepackage{graphicx}
\usepackage{gensymb}

\definecolor{lightdustygreen}{rgb}{0.67, 0.82, 0.69}
\definecolor{lightdustypink}{rgb}{0.87, 0.67, 0.70}
\definecolor{lightdustypurple}{rgb}{0.80, 0.68, 0.83}
\definecolor{dustygray}{rgb}{0.55, 0.55, 0.55}
\definecolor{dustypurple}{rgb}{0.63, 0.50, 0.68}
\definecolor{dustyblue}{rgb}{0.45, 0.65, 0.75}

\title{Pitfalls of Scale: Investigating the Inverse Task of Redefinition in Large Language Models}

\author{
    Elena Stringli, Maria Lymperaiou,  Giorgos Filandrianos, \\
    \textbf{Athanasios Voulodimos, Giorgos Stamou}\\ 
    School of Electrical and Computer Engineering,  AILS Laboratory\\
    National Technical University of Athens \\
    \texttt{\href{mailto:ele191200@gmail.com}{ele191200@gmail.com}, \{\href{mailto:marialymp@islab.ntua.gr}{marialymp}, \href{mailto:geofila@islab.ntua.gr}{geofila}\}@islab.ntua.gr}, \\
    \texttt{\href{mailto:thanosv@mail.ntua.gr}{thanosv@mail.ntua.gr}, \href{mailto:gstam@cs.ntua.gr}{gstam@cs.ntua.gr}}\\
}

\begin{document}
\maketitle
\begin{abstract}
Inverse tasks can uncover potential reasoning gaps as Large Language Models (LLMs) scale up. In this work, we explore the redefinition task, in which we assign alternative values to well-known physical constants and units of measure, prompting LLMs to respond accordingly. Our findings show that not only does model performance degrade with scale, but its false confidence also rises. Moreover, while factors such as prompting strategies or response formatting are influential, they do not preclude LLMs from anchoring to memorized values.
\end{abstract}

\section{Introduction}
The surprising advent of Large Language Models (LLMs)  has greatly sparked the interest in natural language research, demonstrating remarkable results in several linguistic, reasoning and knowledge retrieval tasks \cite{llm-survey}. LLMs are -seemingly- capable of thinking out-of-the-box \cite{giadikiaroglou-etal-2024-puzzle}, preserving factuality of generated claims \cite{wang-etal-2024-factuality} and effectively collaborating in LLM-based multi-agent environments \cite{rasal2024navigatingcomplexityorchestratedproblem}, assimilating human-like traits in thought patterns and even surpassing humans in world-knowledge recall \cite{world-knowledge-llms}. Nevertheless, LLMs remain pattern learners, despite being exposed to years and years of vast documented human knowledge, making the distinction between memorization and genuine capability increasingly ambiguous \cite{wu-etal-2024-reasoning}.

There is evidence that LLMs fall short in truly comprehending human language and cognition in conjunction to its biological imprints on the human brain, as well as its cultural evolution \cite{llm-limits}. This poses a possible inherent divergence between human and LLM reasoning, inspiring the research of breaking points regarding LLM capacity, the more they exhibit advancements in challenging tasks. 
In an effort to formally describe and predict LLM capabilities, \citet{kaplan2020scalinglawsneurallanguage} proposed scaling laws of LLMs, establishing a framework that links model performance to key factors such as parameter count, dataset size, and computational resources. They demonstrate that increasing these variables leads to predictable improvements in language modeling efficiency, shedding light on trade-offs and limitations ingrained in scaling.
\begin{figure}
\vskip -0.08in
    \centering
    \includegraphics[width=0.77\linewidth]{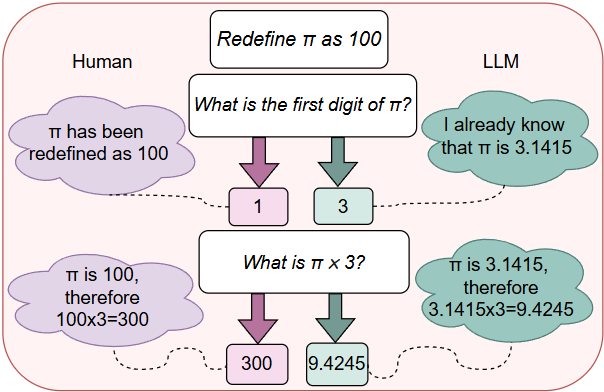}
        \vskip -0.02in
    \caption{Redefined reasoning pathways.}
    \label{fig:redefine-teaser}
    \vskip -0.12in
\end{figure}
Beyond such predictable improvements, larger models often exhibit emergent abilities \cite{wei2022emergentabilitieslargelanguage, Srivastava2023BeyondTI}—capabilities absent in smaller models yet arising spontaneously once a critical scale is reached.  These include in-context learning \cite{in-context}, advanced reasoning \cite{step-by-step}, and compositional generalization \cite{chen-etal-2024-skills}, suggesting that scaling is not merely a linear enhancement of existing skills but also a rather unpredictable threshold mechanism for qualitative shifts in capability.

In an attempt to question emergent abilities as an analogy to model scale, \textit{inverse scaling tasks} \cite{mckenzie2024inversescalingbiggerisnt} re-frame the justified so far trustworthiness that larger models offer. These tasks refer to worsening model performance as the loss on the original training objective improves, contrary the the typical scaling laws that guarantee predictable performance advancements with loss decrease \cite{kaplan2020scalinglawsneurallanguage}. They are designed to expose more potent LLMs, revealing reasoning divergence in comparison to humans, who are able to solve many of these tasks with ease.

Interestingly, inverse scaling is widely underexplored in literature. In this paper, we address this gap by examining the \textit{redefinition task}, where well-known concepts are assigned alternative values, and LLMs are prompted to respond accordingly.  For example, redefining $\pi=100$ (Figure \ref{fig:redefine-teaser}), overwriting the default $\pi=3.14159$ refutes LLM's prior knowledge, calling for flexible reasoning pathways in order to handle mathematical operations over the redefined $\pi$ value.
Through vast experimentation on several redefinitions, LLM families and model scales, we conclude that:
\begin{itemize}
    \item Anchoring to prior knowledge is more prominent in larger models, demonstrating diminishing reasoning flexibility with scale.
    \item Prompting techniques influence anchoring rates but they cannot eliminate the problem.
    \item Larger models prefer to fail than abstain from responding more often than smaller ones.
\end{itemize}

\section{Related work}
\paragraph{Inverse scaling problems} have been thoroughly investigated within the Inverse Scaling Prize contest \cite{mckenzie2024inversescalingbiggerisnt}, targeting to unveil the causes behind inverse scaling. One primary cause is \textit{strong priors}, where the LLM relies on its preexisting knowledge instead of adhering to prompt instructions. Another contributing factor is \textit{unwanted imitation}, where the LLM reproduces undesirable patterns from its training data. Additionally, exemplars containing \textit{distractors} can mislead the LLM by providing easier reasoning shortcuts, obscuring the true task objective. Finally, \textit{spurious few-shot} prompting may steer the LLM toward deceptive reasoning pathways, even when the right answer is explicitly provided in the prompt. Redefinition falls under the category of \textit{strong priors}, achieving 100\% human accuracy—highlighting humans' ability to effortlessly override default meanings. This finding is on par with evidence that given ample time, humans have the cognitive abilities to generalize on alternative realities \cite{wu-etal-2024-reasoning}.
\paragraph{True LLM Reasoning} is a fundamental concern, questioning the real barrier between LLMs and human cognition.  While LLMs excel in linguistic competence, this ability is dissociated with thought \cite{Mahowald2024DissociatingLA}.
In practice, LLMs are prone to performance degradation under alternative formulations, denoting their limited reasoning flexibility \cite{wu-etal-2024-reasoning} and their susceptibility to knowledge conflicts between contextual prompt information and stored facts \cite{xu-etal-2024-knowledge-conflicts}. Similar findings are reported in causal  \cite{jin2024largelanguagemodelsinfer, Gendron2024CanLL}, analogical \cite{Lewis2024UsingCT, Stevenson2024CanLL} and commonsense \cite{Nezhurina2024AliceIW} reasoning, where LLM performance declines sharply under diverging formulations. Alternative prompts are also shown to influence LLM capacity in arithmetic reasoning \cite{Ball2024CanWC, li-etal-2024-challenging}, translation over artificial languages and deductions with twists \cite{li-etal-2024-challenging}. Quite often, memorization accounts for reasoning, perplexing the evaluation of the real LLM abilities \cite{xie2024memorizationlargelanguagemodels, lou2024quantifyingincontextreasoningeffects, wang2024generalizationvsmemorizationtracing}.

\section{Method}
We test redefinition on two kinds of well-encoded knowledge in LLMs. The first one includes widely known physical and mathematical \textbf{constants}, while the second involves commonly used \textbf{units of measure}. We also examine two redefinition types, initially focusing on simple \textit{assignment} of a new value, overriding the default one. A more challenging option is to \textit{swap} two constants/units (e.g. "redefine $\pi$ as $\phi$"), where the LLM has to override its knowledge with another piece of learned information.
Additionally, we design escalating \textit{redefinition levels}, as well as three \textit{question levels}  over original and redefined values, reflecting increasing difficulty.  Finally, from the LLM's response format side, we study both free-form (FF) generation and multiple choice (MC). In the MC case, the problem may become more constrained, but we select distractors that are sufficiently challenging.

\begin{table*}[h!]
\vskip -0.1in
    \centering\small
    \begin{tabular}{p{0.5cm}|cc|ccc|cc}
\hline
 & Actual value & Unit & $R_a 1$ & $R_a 2$ &  $R_a 3$ & $R_s 1$ & $R_s 2$ \\
\hline
$\pi$ & $3.14159$ & - &  $4.5$ & $500$ & $-10$ & $\phi$ & $h$ \\
$e$ & $2.71828$ & - & $9$ & $1300$ & $1.5\times 10^{-12}$ & $pi$ & $k_B$ \\
$\phi$ & $1.61803$ & - & $3.6$ & $321$ & $-2.2$ & $e$ & $N_A$ \\
$c$ & $299,792,458$ & $m/s$ & $2.3\times 10^8$ & $10$ & $-4\times 10^8$ & $N_A$& $q_e$ \\
$G$ & $6.674\times 10^{-11}$ & $m^3/kg*s^2$ & $1.1\times 10^{-10}$ & $50$ & $-525$ & $q_e$ & $pi$ \\
$h$ & $6.626\times 10^{-34}$ &  $J*s$ & $5\times 10^{-33}$ & $482$ & $-0.2$ & $k_B$ & $\phi$ \\
$q_e$ & $1.602\times 10^{-19}$ & $C$& $2.4\times 10^{-21}$ & $3\times 10^4$ & $3\times 10^{50}$ & $\epsilon _0$ & $\pi$ \\
$N_A$ & $6.022\times 10^{23}$ &  $mol^{-1}$ &$8.23\times 10^{23}$ & $75$ & $-1$ & $\overline{R}$ & $e$ \\
$k_B$ & $1.380649\times 10^{-23}$ &  $J/K$ & $4.56\times 10^{-24}$ & $80$ & $-9.9\times 10^{-3}$ & $\epsilon _0$& $pi$ \\
$\overline{R}$ & $8.314$ & $J/(mol*K)$ & $13$ & $3500$ & $-400$ & $\pi$ & $c$ \\
$i$ & $\sqrt{-1}$ & -& $\sqrt{-2}$ & $\sqrt{-100}$ & $1$ & $\phi$ & $\overline{R}$ \\
$\sqrt{2}$ & $1.41421356$ & -& $5$ & $31.62$ & $-2$ & $\pi$ & $\epsilon _0$ \\
$\infty$ & infinity has no value & -& $10^{10}$ & $100$ & $-1$ & $c$ &$q_e$\\
$\epsilon _0$ & $8.854\times 10^{-12}$ & $F/m$ & $9.3\times 10^{-10}$ & $35$ & $3\times 10^{12}$ & $G$ & $\phi$ \\
zero & $0$ & -& $-1$ & $100$ & $5\times 10^{30}$ & $h$ & $c$ \\
\hline
    \end{tabular}
    \vskip -0.01in
    \caption{Varying levels of difficulty for constant redefinitions (assignments and swaps).}
    \label{tab:constants-redefine}
        \vskip -0.01in
\end{table*}
\paragraph{Constants redefinition}
involves the following: $\pi$, Euler's number $e$, $\phi$, the speed of light $c$, the gravitational constant $G$, Planck's constant $h$, the elementary charge $q_e$,  Avogadro's number $N_A$, the Boltzmann constant $k_B$, the gas constant $\overline{R}$, the imaginary $i$, the square root of 2 ($\sqrt{2}$), infinity $\infty$,
the vacuum electricity permittivity $\epsilon _0$ and \textit{zero}.
\begin{table}[t!]
\vskip -0.08in
    \centering\small
    \begin{tabular}{p{0.4cm}|p{6.6cm}}
\hline
 & \textbf{$Q_2$} \\\hline
$\pi$ & What is $\pi$ multiplied by 3? \\
$e$ & What is $e^2?$\\
$\phi$ & What is $5*\phi-2$? \\
$c$ & How much time (in sec) does it take light to travel a distance of 100 million km? \\
$G$ & What the gravitational constant multiplied by 7? \\
$h$ & If the frequency of a photon is 4 Hz, what is its energy? Use the formula $E=h*v$. \\
$q_e$ & If an electron has a charge of $-e$, what is the charge of two electrons? \\
$N_A$ & How many atoms are there in $1 mol$ of any element? \\
$k_B$ & Calculate the energy associated with a temperature of 300 K for a  particle using the formula $E=kT$. \\
$\overline{R}$ & What is the gas constant divided by 2? \\
$i$ & What is the value of $i^3$? \\
$\sqrt{2}$ & Calculate the value of squared root of 2 multiplied by 3. What is it approximately? \\
$\infty$ & What is the limit of $1/x$ as $x$ approaches infinity? \\
$\epsilon_0$ & If you add the value of vacuum electric permittivity to itself, what do you get? \\
zero & What is $300$ multiplied by zero? \\
\hline
    \end{tabular}
    \vskip -0.01in
    \caption{$Q2$ questions per constant.}
    \label{tab:q2}
    \vskip -0.12in
\end{table}

\begin{table}[h!]
\vskip -0.08in
    \centering\small
    \begin{tabular}{p{0.4cm}|p{6.6cm}}
\hline
 & \textbf{$Q_3$} \\\hline
$\pi$ & What is the Earth's surface area? \\
$e$ & If a population grows continuously at a rate of 5\% per year, by what factor will it increase in 10 years? \\
$\phi$ & If a rectangle has sides in the golden ratio and the longer side is 8 cm, what's the length of the other side? \\
$c$ & What is the energy equivalent of 8 grams of mass? \\
$G$ & If two $15$ kg masses are placed 2 meters apart, calculate the gravitational force between them. \\
$h$ & In the photoelectric effect, if a metal has a work function of $4.5\times 10^{-19}J$, what is the minimum frequency of light required to eject an electron from the metal surface? \\
$q_e$ & A capacitor stores a charge of $3.2\times 10^{-18}$ coulombs. How many elementary charges $e$ are equivalent to this amount of charge? \\
$N_A$ & Calculate the number of molecules in $54 grams$ of water (molar mass of water is  $\sim 18 g/mol$). \\
$k_B$ & What is the temperature at which the average kinetic energy of a particle is $1.9\times 10^{-21}J$? \\
$\overline{R}$ & If you have 2 moles of an ideal gas at a temperature of $300 K$, what is the pressure (in $Pa$) if the volume is $10 liters$? \\
$i$ & If $z_1=1+i$ and $z_2=1-i$, calculate  $z_1\cdot z_2$. \\
$\sqrt{2}$ & If one side of a square is 5 units long, what is the length of the diagonal of the square? \\
$\infty$ & What is the horizontal asymptote of the function $f(x) = (5x+30000)/(x+1000), x>0$? \\
$\epsilon_0$ & Calculate the electric force between two charges $q_1=3\mu C$ and $q_2=5\mu C$ separated by 12m in a vacuum. \\
zero & If $y = \sin(x)/x$, what is the limit of $y$ as $x$ approaches 0? \\
\hline
    \end{tabular}
    \vskip -0.02in
    \caption{$Q3$ questions per constant.}
    \label{tab:q3}
        \vskip -0.13in
\end{table}
\begin{table*}[h!]
\vskip -0.05in
    \centering\small
    \begin{tabular}{p{0.9cm}|cc|ccc}
\hline
Unit & Derived unit & Actual value & $R_a 1$ & $R_a 2$ & $R_a 3$ \\
\hline
1 \textit{min} & seconds (\textit{sec}) & $60 \textit{sec}$ & $100 \textit{sec}$ & $5\times 10^8\textit{sec}$ & $-50 sec$ \\
1 \textit{kg} & grams (\textit{gr}) & $1000 \textit{gr}$ & $900 \textit{gr}$ & $10^{-14} \textit{gr}$ & $-100 \textit{gr}$ \\
1 \textit{ m} & centimeter (\textit{cm}) & $100 cm$ & $60 cm$ & $3×10^10 cm$ & $-200 cm$ \\
\textit{K} & Celsius degrees (\degree \textit{C})& $\degree C + 273.15$ & $\degree C + 300$ & $\degree C + 1$ & $100*\degree C + 500$ \\
1 \textit{mL} & cubic centimeter ($cm^3$) & $1 cm^3$ & $2 cm^3$ & $10000 cm^3$ & $-10 cm^3$ \\
1 \textit{cal} & Joule (\textit{J}) & $4.184 J$ & $9 J$ & $1500 J$ & $-5 J$ \\
1 \textit{atm} & Pascal (\textit{Pa}) & $101,325 Pa$ & $215,000 Pa$ & $0.55 Pa$ & $-5000 Pa$ \\
1 \textit{V} & milivolt (\textit{mV}) & $1000 mV$ & $500 mV$ & $4×10^9 mV$ & $-10 mV$ \\
1 \textit{MHz} & Hertz (\textit{Hz}) & $10^6 Hz$ & $10^5 Hz$ & $2 Hz$ & $-10^3 Hz$ \\
1 \textit{N} & millinewton (\textit{mN}) & $1000 mN$ & $900 mN$ & $2×10^15 mN$ & $-3000 mN$ \\
1 \textit{kW} & Watt (\textit{W}) & $1000 W$ & $1500 W$ & $5×10^{-5} W$ & $-30 W$ \\
1 \textit{T} & millitesla (\textit{mT}) & $1000 mT$ & $600 mT$ & $10^23 mT$ & $-90 mT$ \\
1 \textit{ha} & square meter ($m^2$)& $10,000 m^2$ & $10,500 m^2$ & $3×10^{-4} m^2$ & $-25 m^2$ \\
1 \textit{lx} & lumen per $m^2$ ($lm/m^2)$ & $1 lm/m^2$ & $0.5 lm/m^2$ & $1000 lm/m^2$ & $-19 lm/m^2$ \\
1 \textit{ly} & Trillion/Billion \textit{km} & $9.461 T km$ & $9.461 B km$ & $10 m$ & $-2 T km$ \\
1 \textit{B} & bit (\textit{b}) & $8 b$ & $10 b$ & $6×10^8 b$ & $-4 b$ \\
\hline
    \end{tabular}    
    \vskip -0.01in
    \caption{Redefinitions of unit scaling between base and derived units.}
 \label{tab:redefinition-units}
        \vskip -0.1in
\end{table*}
We then design \textit{assignment} redefinitions $R_a$ for the three degrees of increasing difficulty. In the first level, we assign a value close to the actual one ("redefine $\pi$ as $4.5$"), inspecting how an LLM handles variance within an acceptable range. To stress the LLM's flexibility, we modify values by orders of magnitude, assigning a deviating value ("redefine $\pi$ as $500$") in the second level. In the third level, we move to unrealistic values, assigning negative numbers to constants ("redefine $\pi$ as $-10$"). In the \textit{swapping} case ($R_s$), we impose two difficulty levels, with the first one concerning values close to the actual (e.g. "redefine $\pi$ as $\phi$", since the actual values of $\pi=3.14159$ and $\phi=2.71828$ are close), while the second level imposes swapping of constants differing by orders of magnitude (e.g. "redefine $\pi$ as the \textit{Planck's constant}", where  Planck's constant=$6.626\times 10^{-34}$).
All constant redefinitions are presented in Table \ref{tab:constants-redefine}.

We also design three levels of question difficulty. The first level  ($Q_1$) mainly regards the question \textit{What is the first -non-zero- digit of \{constant\}?}. The correct answer $A_{Q_1}$ is actually isolating the leftmost digit (ignoring leading zeros or the minus sign in cases of negative numbers) of the constant. For example, when $\pi$ has undergone the redefinition $\pi=500$ the correct response $A_{Q_1}$ is 5. There are some exceptions to the first digit $Q_1$, (presented in App. \ref{sec:exceptions-q1}).
The next question level ($Q_2$), asks for a simple mathematical operation (e.g. \textit{What is $\pi$ multiplied by 3?}), as presented in Table \ref{tab:q2}. The LLM has to execute this operation correctly to derive the correct $A_{Q_2}$, while the ground truth solution can be reached by utilizing a scientific calculator and the appropriate constant value. Finally, in the last and most difficult level ($Q_3$), questions  requiring multi-hop reasoning are designed (e.g. \textit{What is the Earth's surface area?}),  as the ones of Table \ref{tab:q3}.
\paragraph{Units of measure redefinition}
incorporates the following fundamental physical quantities: time (minutes-\textbf{\textit{min}}), weight (kilogram-\textbf{\textit{kg}}), length (meter-\textbf{\textit{m}}) and light-year (\textbf{\textit{ly}}), temperature (Kelvin-\textbf{\textit{K}}), volume (milliliter-\textbf{\textit{mL}}), energy (calorie-\textbf{\textit{cal}}), pressure (atmosphere-\textbf{\textit{atm}}), voltage (Volt-\textbf{\textit{V}}), frequency (megaHz-\textbf{\textit{MHz}}), force (newton-\textbf{\textit{N}}), magnetic flux density (Tesla-\textbf{\textit{T}}), area (hectare-\textbf{\textit{ha}}), illuminance (lux-\textbf{\textit{lx}}), and information (byte-\textbf{\textit{B}}). We intervene on the scaling between each of those units and their derived counterparts for the same physical quantity: for example, a minute has 60 seconds, therefore a unit redefinition can be "redefine minutes to have 100 seconds". Details about such redefinitions are presented in Table \ref{tab:redefinition-units}.

As in the constants' case, we offer three levels of questions difficulty. The easiest $Q_1$ level queries the actual conversion rule as defined in Physics, with a small adjustment to avoid the trivial case, where the answer lies in the prompt: instead of questioning \textit{How many seconds a minute is?}, since its actual rephrasing exists in the prompt ("redefine a minute to have 100 seconds"), we prefer questions such as \textit{How many seconds are in two minutes?}, imposing an undemanding calculation. In the $Q_2$ case, the LLM is tasked to solve an easy problem, applying fundamental physics equations or a unit scaling given minimal context. In the hardest $Q_3$ level, questions require more mathematical reasoning steps. All questions are illustrated in Table \ref{tab:questions-units}.
\begin{table*}[h!]
\vskip -0.05in
    \centering\small
    \begin{tabular}{p{0.3cm}p{2.5cm}p{4.2cm}p{7.5cm}}
\hline
& $Q_1$ & $Q_2$ & $Q_3$ \\
\hline
\textit{min} & How many \textit{sec} are in 2 \textit{min}? & A stopwatch runs for 3 and a half \textit{min}. How many sec does it count? & A marathon runner runs at a speed of 170 \textit{m/min}. How many \textit{sec} will it take them to complete a 42-\textit{km} race? \\
\textit{kg} & How many \textit{gr} are in 2 \textit{kg}? & A person weighs 72 \textit{kg}. What is the persons weight in \textit{gr}? & A vehicle's engine weighs 650 \textit{kg}. If 15\% of the weight is aluminum, what is the weight of the aluminum in \textit{gr}? \\
\textit{m} & How many \textit{cm} are in 2 \textit{m}? & A circular track has a circumference of 400 \textit{m}. What is its diameter in \textit{cm}? & If a rectangular field is 50 \textit{m} long and 30 \textit{m} wide, what is its area in $cm^2$? \\
$K$ & What is the $K$ temperature when it is 0°C? & Water boils at 100°C. What is its boiling point in \textit{K}? & At a certain point in time, the temperature of a black hole's event horizon is measured to be 20°C. If the temperature in °C decreases by 30\% after an event, what is the new temperature in \textit{K}? \\
\textit{mL} & How many \textit{mL} are in 1 $cm^3$? & If you have a container that holds 1,250 \textit{mL} of liquid, how many $cm^3$ of liquid can it hold? & A spherical ball has a radius of 10 \textit{cm}. What is its volume in \textit{mL}? \\
\textit{cal} & How many $J$ are in 3 \textit{cal}? & A person burns 200 $J$ of energy while jogging. How many \textit{cal} did they burn? & A car burns 3,400 $J$ of fuel every \textit{min}. If the car runs for 2 hours, how many \textit{cal} does it burn? \\
\textit{atm} & How many \textit{Pa} are in 2 \textit{atm}? & A diver is 100 $m$ below the surface of the ocean where the pressure is 152,300 \textit{Pa}. How many \textit{atm} of pressure are they experiencing? & A pressurized gas tank holds a gas at a pressure of 150,000 \textit{Pa}. If the gas occupies a volume of 4 $m^3$ at this pressure, and the gas is suddenly released to 2 \textit{atm}, what will be the new volume of the gas? Assume temperature and the number of gas molecules remain constant and use Boyle's Law. \\
$V$& How many \textit{mV} are in 5 $V$? & A circuit is powered by 30,000 \textit{mV}. How many $V$is this? & A battery supplies 100,000 $mV$ to a device. If the device operates with a resistance of 20 ohms, what is the current (in Amperes) flowing through the device using Ohm's Law? \\
\textit{MHz} & How many \textit{Hz} are in 2 \textit{MHz}? & An oscillator operates at 4 \textit{MHz}. What is the period of the wave in \textit{sec}? & A circuit has a signal with a frequency of 6 \textit{MHz}. What is the wavelength of the signal if the speed of light is approximately $3\times10^8$ \textit{m/s}? \\
\textit{N} & How many \textit{mN} are in 2 \textit{N}? & A person applies a force of 24 \textit{N} to a cart with a mass of 3 \textit{kg}. What is the is the force applied to the cart by the person in \textit{mN}? & A 10-\textit{kg} object is pulled with a force of 4,300 \textit{mN}. What is the acceleration of the object ($m/s^2$)? \\
\textit{kW} & How many W are in 2 \textit{kW}? & A lightbulb consumes 900 \textit{W} of power. How many \textit{kW} is this? & A factory uses 12 \textit{kW} for 10 hours per day for 30 days. What is the total energy consumption in watt-hours? \\
\textit{T} & How many \textit{mT} are in 3 \textit{T}? & A coil generates a magnetic field of 300 \textit{mT}. What is this field strength in \textit{T}? & A particle moves through a magnetic field of 3,600 \textit{mT} with a charge of $2\times 10^{-6}$ $C$ and a velocity of $10^5$ $m/s$. What is the magnetic force on the particle? \\
ha & What is the area of 2 $ha$ in $m^2$? & A park has an area of 86,000 $m^2$. How many $ha$ is the park? & A triangular plot of land has a base of 300 \textit{m} and a height of 350 \textit{m}. How many $ha$ is the plot? \\
\textit{lx} & How many \textit{lx} are equivalent to 4 $lm/m^2$? & A workspace is illuminated at a level of 6 \textit{lx}. What is the illumination in $lm/m^2$? & A light source emits 300 \textit{lm} uniformly over a circular area with a radius of 10 $m$. What is the average illumination in $lx$ over this area? \\
$ly$ & How many \textit{km} are in 2 $ly$? & The Andromeda Galaxy is approximately 23 $ly$ from Earth. What is this distance in \textit{km}? & A black hole is 150 $ly$ away. If light travels at a speed of 0.3 billion $km/s$, how long would it take for light to travel this distance in $sec$? \\
$B$ & How many $b$ are in 3 $B$? & If a document is 8,000 $b$ in size, how many $B$ does it occupy? & A 1-\textit{min} high-definition video uses a data rate of $8\times 10^6$ $B/sec$. How many $b$ does the video consume in total? \\
\hline
    \end{tabular}
    \caption{Questions of three difficulty levels ($Q_1$, $Q_2$, $Q_3$) for units of measure.}
    \label{tab:questions-units}
        \vskip -0.08in
\end{table*}

\section{Experiments}
We test 19 LLMs, including state-of-the-art (SoTA) model families: Llama 3 (8/70/405B), Mistral7B/Large/Mixtral8$\times$7b, Anthropic Claude (Opus/Instant/Haiku/v2/Sonnet 3.5\&3.7), Cohere command (light/text/r/r+) and Amazon Titan (text lite/text express/large). 
All LLMs are prompted using zero shot (ZS), few shot (FS) and Chain of Thought (CoT) techniques. 

For evaluation, we decompose the LLMs' responses, assigning them to four categories:

\textbf{1. No redefinition (NR) correct responses:} These correspond to cases that the LLM indeed knows the response correctly before redefinition.

\textbf{2. Anchored responses:} These were correct before redefinition, but incorrect afterwards, e.g. replying that 3 is the first digit of redefined $\pi=100$ reveals an excessive anchoring to prior knowledge.

\textbf{3. Correct responses:} The LLM fully adopts the redefined concept and responds accordingly.

\textbf{4. Completely wrong responses:} The LLM produces blank, incorrect or inconsistent responses that do not fit any of the above cases. In some cases, it completely refuses to perform the redefinition.

To measure the impact of redefinitions, results post-redefinition are compared with those where no redefinition is performed (denoted as \textbf{NR}). We then focus on \textbf{anchored responses}, since they are mostly tied to the memorization versus reasoning trade-off in LLMs.

\subsection{Results on constants redefinition}
An overview of response accuracy is presented in Table \ref{tab:anchored_table}, where we consider the hardest redefinitions ($R_a3$ and $R_s2$ for \textit{assignment} and \textit{swapping} respectively), as well as all three question levels, together with FF and MC response formats. It is observable that all tested LLMs, regardless of their size or model family, are prone to anchoring. This is especially evident in the FF format (since MC introduces a random choice factor), where models such as Titan Large generate 60\% anchored responses, while Claude Opus and Command r produce 47\% and 53\% respectively in this format.

\begin{table*}[h!]
\vskip -0.08in
\centering
\small
\begin{tabular}{p{2.7cm}|p{0.7cm}p{0.6cm}|p{0.7cm}p{0.6cm}|p{0.7cm}p{0.6cm}|p{0.7cm}p{0.6cm}|p{0.7cm}p{0.6cm}|p{0.7cm}p{0.6cm}}
\hline
\multirow{3}{*}{Model} & \multicolumn{6}{c|}{$R_a3$}                                                               & \multicolumn{6}{c}{$R_s2$}                                                                \\ \cline{2-13}
                       & \multicolumn{2}{c}{$Q_1$} & \multicolumn{2}{c}{$Q_2$} & \multicolumn{2}{c|}{$Q_3$} & \multicolumn{2}{c}{$Q_1$} & \multicolumn{2}{c}{$Q_2$} & \multicolumn{2}{c}{$Q_3$} \\ \cline{2-13}
                       & FF           & MC           & FF           & MC           & FF            & MC           & FF           & MC           & FF           & MC           & FF           & MC           \\ \hline

Mistral7B & \textbf{33.33} & \textbf{46.67} & \textbf{33.33} & \textbf{26.67} & 26.67 & 40.0 & 33.33 & 53.33 & 13.33 & 33.33 & 26.67 & 20.0 \\ 
Mixtral8x7B & 33.33 & 33.33 & 26.67 & 26.67 & 20.0 & 33.33 & 26.67 & 46.67 & 40.0 & \textbf{53.33} & 46.67 & \textbf{73.33} \\ 
Mistral Large (123B) & 33.33 & 20.0 & 26.67 & 26.67 & \textbf{53.33} & \textbf{66.67} & \textbf{66.67} & \textbf{53.33} & \textbf{46.67} & 40.0 & \textbf{73.33} & 66.67 \\  \hline
Llama8B & 0.0 & \textbf{26.67} & 0.0 & \textbf{26.67} & 13.33 & 33.33 & 20.0 & 13.33 & \textbf{26.67} & 40.0 & 20.0 & 20.0 \\ 
Llama70B & \textbf{6.67 }& 13.33 & 0.0 & 0.0 & 13.33 & 40.0 & 33.33 & 46.67 & 13.33 & \textbf{46.67} & 33.33 & 73.33 \\ 
Llama405B & 0.0 & 0.0 & 0.0 & 13.33 & \textbf{26.67} & \textbf{53.33} & \textbf{26.67} & \textbf{46.67} & 6.67 & 20.0 & \textbf{53.33} & \textbf{93.33} \\  \hline
Titan lite    & 13.33 & 20.0  & 20.0  & 20.0  & 0.0   & 40.0  & 40.0  & 33.33  & 20.0  & 33.33  & 6.67   & 26.67 \\ 
Titan express & 20.0  & \textbf{26.67} & 13.33  & 13.33  & \textbf{20.0}  & 13.33  & 40.0  & \textbf{53.33}  & \textbf{20.0}  & 20.0  & 33.33  & \textbf{26.67} \\ 
Titan large   & \textbf{26.67}  & 20.0  & \textbf{20.0}  & 6.67  & 13.33  & \textbf{40.0}  & \textbf{60.0}  & 40.0  & 13.33  & \textbf{33.33}  & \textbf{33.33}  & 20.0 \\   \hline
Command r & 0.0 & 6.67 & \textbf{20.0} & \textbf{33.33} & \textbf{26.67} & \textbf{53.33} & \textbf{53.33} & 13.33 & 20.0 & 6.67 & \textbf{33.33} & \textbf{46.67} \\ 
Command r + & 6.67 & 13.33 & 0.0 & 13.33 & 13.33 & 26.67 & 13.33 & 20.0 & 26.67 & 6.67 & 33.33 & 26.67 \\ 
Command light text & 6.67 & 13.33 & 13.33 & 20.0 & 0.0 & 40.0 & 13.33 & 20.0 & \textbf{26.67} & \textbf{20.0} & 13.33 & 13.33 \\ 
Command text & \textbf{13.33} & \textbf{20.0} & 6.67 & 6.67 & 6.67 & 26.67 & 40.0 & \textbf{26.67} & 13.33 & 26.67 & 13.33 & 33.33 \\   \hline
Claude Opus & 13.33 & 0.0 & 6.67 & 6.67 & 33.33 & \textbf{46.67} & \textbf{46.67} & \textbf{40.0} & 20.0 & \textbf{26.67} & 53.33 & 73.33 \\ 
Claude Instant & 0.0 & 13.33 & 13.33 & \textbf{20.0 } & 26.67 & 46.67 & 33.33 & 20.0 & 33.33 & 40.0 & 46.67 & 60.0 \\ 
Claude Haiku & 20.0 & 13.33 & 6.67 & 0.0 & 20.0 & 20.0 & 26.67 & 6.67 & 20.0 & 20.0 & 40.0 & 53.33 \\ 
Claude v2 & 26.67  & 13.33  & \textbf{20.0 } & 0.0 & \textbf{46.67} & 40.0 & 13.33 & 40.0 & \textbf{33.33} & 20.0 & 40.0 & 66.67 \\
Claude 3.5 Sonnet& \textbf{26.67} & \textbf{13.33} & 0.0 & 13.33 & 13.33 & 33.33 & 33.33 & 40.0 & 20.0 & 20.0 & \textbf{60.0 } & \textbf{73.33} \\ 
Claude 3.7 Sonnet\tablefootnote{Without thinking module enabled for fair comparison.} & 0.0  & 0.0  & 0.0 & 6.67 & 13.33 & 13.33 & 33.33 & 20.0 & 6.67 & 20.0 & 40.0 & 33.33 \\ 
\hline
\end{tabular}

\caption{Anchoring response rate for all LLMs tested using ZS prompting for the most difficult in \textit{assignment} ($R_a3$) and \textit{swapping} ($R_s2$) redefinitions. The highest anchoring rate for each LLM family is marked in \textbf{bold}.}
\label{tab:anchored_table}
\end{table*}


\begin{table}[h]
\vskip -0.03in
\small
    \centering
    \begin{tabular}{l|p{0.85cm}p{0.85cm}p{0.85cm}|p{0.85cm}p{0.85cm}}
        \hline
        Level & $R_a1$ & $R_a2$ & $R_a3$ & $R_s1$ & $R_s2$ \\ \hline
        \multicolumn{6}{c}{Free-Form (FF)} \\ 
        \hline

        $Q_1$ & \cellcolor{lightdustygreen} -0.458 & -0.071 & 0.008 & 0.199 & -0.016 \\
        $Q_2$ &\cellcolor{lightdustygreen}  -0.502 & \cellcolor{lightdustygreen} -0.573 & \cellcolor{lightdustygreen} -0.472 & 0.107 & 0.019 \\
        $Q_3$ & \cellcolor{lightdustypink} 0.489 & 0.237 & 0.292 & \cellcolor{lightdustypink} 0.666 & \cellcolor{lightdustypink} 0.668 \\ \hline

        \multicolumn{6}{c}{Multiple Choice (MC)} \\ 
        \hline
        $Q_1$ &  \cellcolor{lightdustygreen} -0.642 &  \cellcolor{lightdustygreen} -0.4 &  \cellcolor{lightdustygreen} -0.344 & -0.052 & 0.025 \\ 
        $Q_2$ & -0.275 &  \cellcolor{lightdustygreen} -0.316 & -0.245 & \cellcolor{lightdustypink} 0.41 & 0.151 \\
        $Q_3$ & -0.063 & \cellcolor{lightdustypink} 0.457 & 0.081 & \cellcolor{lightdustypink} 0.666 & \cellcolor{lightdustypink} 0.75 \\ 
        \hline
    \end{tabular}
    \caption{Correlation between average NR correct response rate with anchored response rate for each redefinition and question level in ZS setup. 
    Cells in \textcolor{lightdustypink}{pink} indicate a \textbf{high positive correlation} ($>0.3$), while cells in \textcolor{lightdustygreen}{green} indicate a \textbf{high negative correlation} ($<-0.3$).}
    \label{tab:correlation}
        \vskip -0.12in
\end{table}

To investigate the possible source of this phenomenon, we calculate the correlation between NR rate and post-redefinition anchored responses per LLM. Averaged results for all LLMs are presented in 
Table \ref{tab:correlation}, revealing an intriguing pattern: for $Q_1$ and $Q_2$ levels, the correlation is either weak or negative. A negative correlation indicates that LLMs performing well in NR cases tend to anchor less. This suggests that when reasoning is of easy or medium level, LLMs are less likely to adhere rigidly to their default knowledge. This serves as a sanity check, confirming that LLMs understand the redefinition task and that anchoring rates are not due to prompting deficiencies. 
However, this trend reverses in the most challenging $Q_3$ level, particularly in the \textit{swapping} cases. In these instances, a strong positive correlation is evident, implying that LLMs that originally perform well (thus are potent reasoners) tend to disregard the redefinition prompt and respond as they would in the NR case. This striking observation suggests that \textit{more capable reasoners anchor more on their prior knowledge}.
This pattern holds across both FF and MC formats, as well as different prompt types. More results are provided in Appendix \ref{app:correlation}.

\begin{table*}[h]
\vskip -0.08in
\centering
\small
\begin{tabular}{p{2.7cm}|p{0.7cm}p{0.6cm}|p{0.7cm}p{0.6cm}|p{0.7cm}p{0.6cm}|p{0.7cm}p{0.6cm}|p{0.7cm}p{0.6cm}|p{0.7cm}p{0.6cm}}
\hline
\multirow{3}{*}{Model} & \multicolumn{6}{c|}{$R_a3$}                                                               & \multicolumn{6}{c}{$R_s2$}                                                                \\ \cline{2-13}
                       & \multicolumn{2}{c}{$Q_1$} & \multicolumn{2}{c}{$Q_2$} & \multicolumn{2}{c|}{$Q_3$} & \multicolumn{2}{c}{$Q_1$} & \multicolumn{2}{c}{$Q_2$} & \multicolumn{2}{c}{$Q_3$} \\ \cline{2-13}
                       & NR           & FF           & NR           & FF           & NR            & FF           & NR           & FF           & NR           & FF           & NR           & FF           \\ \hline

Mistral7B & 66.67 & 33.33 & 46.67 & 33.33 & 33.33 & 26.67 & 66.67 & 33.33 & 46.67 & \cellcolor{lightdustypurple}13.33 & 33.33 & \cellcolor{lightdustypurple}26.67 \\ 
Mixtral8x7B & 100.0 & 33.33 & 66.67 & 26.67 & 66.67 & 20.0 & 100.0 & 26.67 & 66.67 & \cellcolor{lightdustypurple}40.0 & 66.67 & \cellcolor{lightdustypurple}46.67 \\ 
Mistral Large (123B) & 93.33 & 33.33 & 73.33 & 26.67 & 53.33 & 53.33 & 93.33 & 66.67 & 73.33 & \cellcolor{lightdustypurple}46.67 & 53.33 & \cellcolor{lightdustypurple}73.33 \\  \hline
Llama8B & 80.0 & 0.0 & 80.0 & 0.0 &  53.33 & \cellcolor{lightdustypurple}13.33 & 80.0 & 20.0 & 80.0 & 26.67 & 53.33 & \cellcolor{lightdustypurple}20.0 \\ 
Llama70B & 93.33 & 6.67 & 80.0 & 0.0 & 80.0 & \cellcolor{lightdustypurple}13.33 & 93.33 & 33.33 & 80.0 & 13.33 & 80.0 & \cellcolor{lightdustypurple}33.33 \\ 
Llama405B & 93.33 & 0.0 & 86.67 & 0.0 & 73.33 & \cellcolor{lightdustypurple}26.67 & 93.33 & 26.67 & 86.67 & 6.67 & 73.33 & \cellcolor{lightdustypurple}53.33 \\  \hline
\end{tabular}
\vskip -0.01in
\caption{Correct response rate without redefinition (NR) versus post-redefinition anchoring rate in the free-form (FF) format, for LLMs with known sizes using ZS prompting. Colored cells indicate elevated anchoring with LLM scale.}
\label{tab:anchored_table_NR}
\end{table*}

\begin{figure} [t!]
\vskip -0.01in
\centering 
\includegraphics[width=0.78\linewidth]{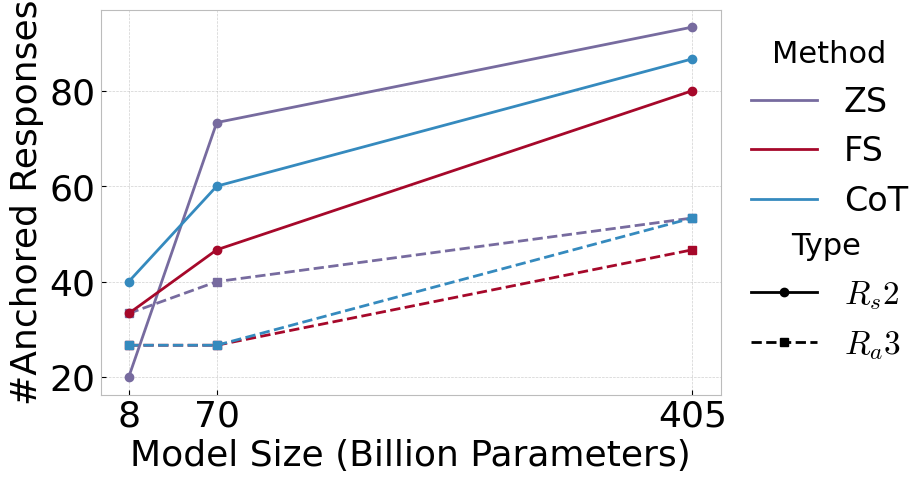}
\caption{Number of anchored responses for models of varying sizes in the Llama family (MC response format).} \label{fig:model_size_vs_anchored_responses} 
\vskip -0.09in
\end{figure}

\paragraph{Inverse trends}
Anchoring is not only related to the per LLM reasoning capabilities, but also to the parameter size of the LLM itself.
Even though larger models achieve higher correct response rate across redefinition levels, staying on par with their reasoning capabilities in the NR case, \textit{the anchoring rate also rises as LLM size increases}. This indicates that larger models struggle to redefine well-known concepts, and instead rely on their existing knowledge. This is evident from Table \ref{tab:anchored_table_NR}, which shows the number of correct responses in the NR case versus the anchoring rate post-redefinition. For example, in the case of Llama, the 405B model anchors significantly when solving $Q_3$ questions post-redefinition compared to the smaller Llama70B, suggesting that the larger variant is less capable as a reasoner. The same trend holds for Mixtral8x7B and Mistral Large (123B): for the latter,  anchoring is even higher compared to correct NR responses in the $Q_3$ level, meaning that the LLM provides the originally correct answer in the redefined problem (when this response is \textit{incorrect}) more frequently than in the NR case (when the answer is \textit{correct}).
This unexpected behavior of Llama is further investigated under the MC response format for different prompting methods, focusing on the hardest redefinition ($R_a3$, $R_s2$) and  question levels ($Q_3$). As illustrated in Figure \ref{fig:model_size_vs_anchored_responses}, the anchoring rate rises with model scale, verifying the \textit{inverse scaling trend}.
\begin{figure*}[h]
\begin{subfigure}{\textwidth}
        \centering
         \includegraphics[width=0.86\linewidth]{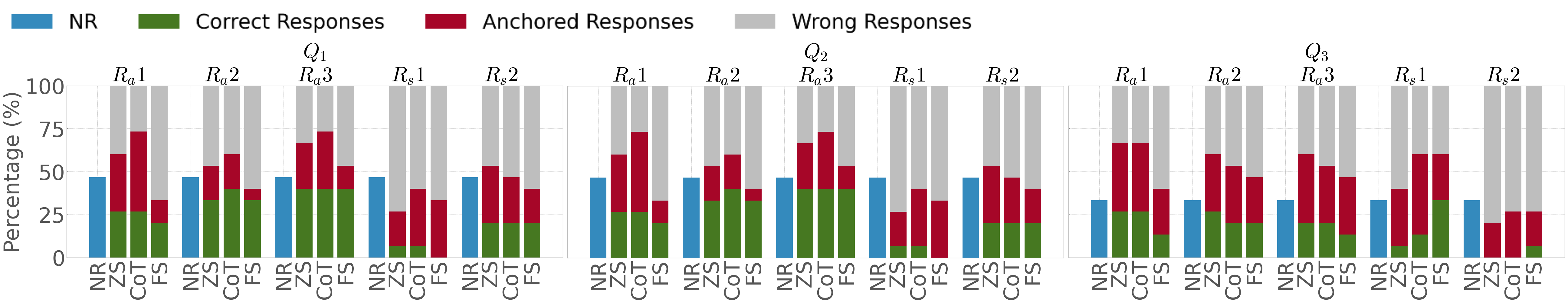}
         \vskip -0.01in
        \caption{Response breakdown for Mistral 7B before and after constant redefinitions.}
        \label{fig:mistral_large_MC}
    \end{subfigure}
    
    \begin{subfigure}{\textwidth}
        \centering
        \includegraphics[width=0.86\linewidth]{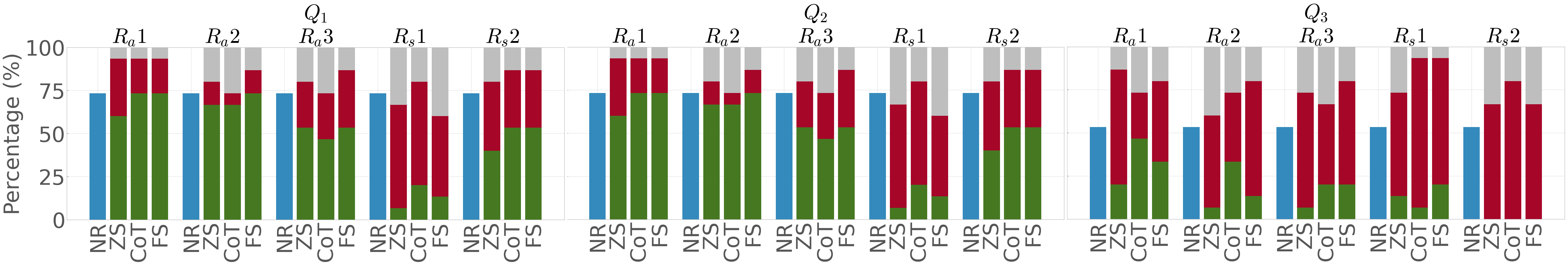}
        \vskip -0.01in
        \caption{Response breakdown for Mistral Large (123B) before and after constant redefinitions.}
        \label{fig:mistral_large_MC}
    \end{subfigure}
    \vskip -0.01in
    \caption{Comparison of Mistral 7B and Mistral Large  responses on the MC response format.}
    \label{fig:mistral_all}
\end{figure*}
The same holds for Mistral, as shown in Figure \ref{fig:mistral_all}, which illustrates the performance of Mistral 7B and Large. Once again, the larger model consistently exhibits a much higher anchoring rate, regardless of the redefinition type or the difficulty of the question—sometimes even exceeding its performance in the NR case.



\begin{figure*}[h!]
\vskip -0.01in
    \centering
    \subfloat[Response breakdown for Mistral models.]{ 
        \includegraphics[width=0.325\linewidth]{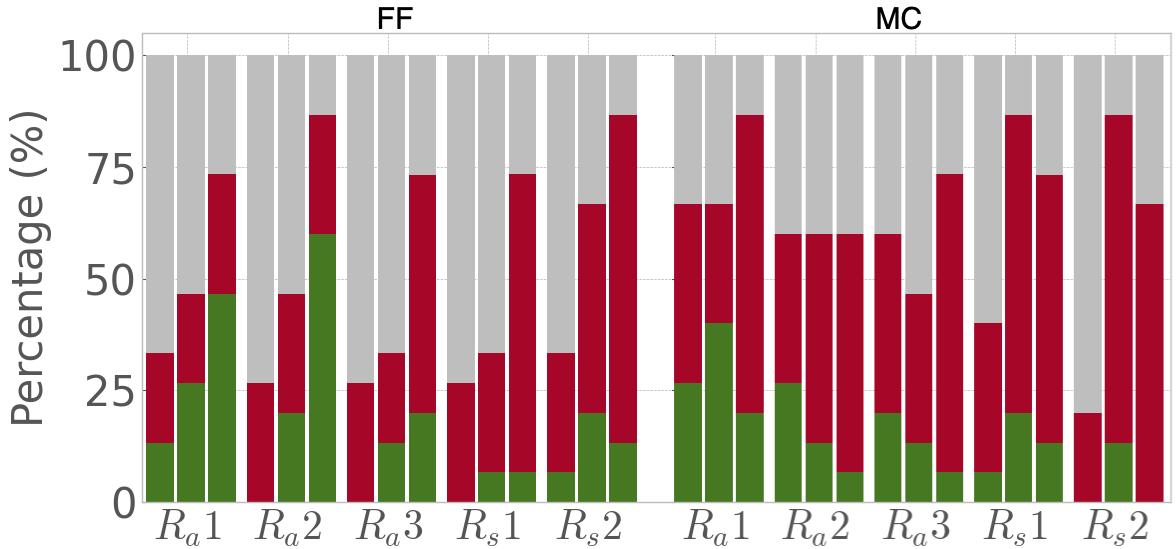}
        \label{fig:mistral}
    }   \hspace{0.5cm}
    \subfloat[Response breakdown for Llama models.]{ 
        \includegraphics[width=0.325\linewidth]{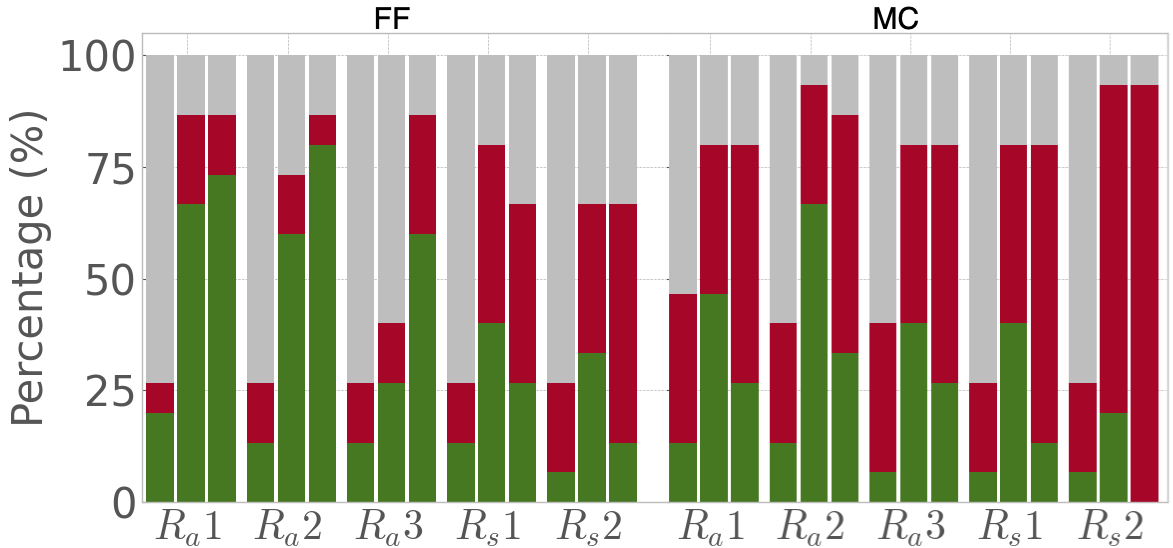}
        \label{fig:llama}
    } \\
        \vskip -0.01in
    \caption{Results for the different Mistral and Llama models on $Q_3$ questions using ZS prompting. The order of the bars per redefinition type/level corresponds to increasing model size. The color coding is the same as in Figure \ref{fig:mistral_all}.}
    \label{fig:size_comparison}
    \vskip -0.09in
\end{figure*}

\paragraph{Response format}
Figure \ref{fig:size_comparison} presents  results from two LLM families of known parameter count, Mistral and Llama, with varying sizes, for all redefinition levels on  $Q_3$ questions, for FF and MC formats. The MC format is associated with higher anchoring rates (e.g., 73.33\% and 93.33\% for Llama 70B and 405B respectively) compared to FF responses (33.33\% and 53.33\%). This is rather expected, since the LLM is exposed to the default value of a constant in the presence of the correct MC candidate, creating a conflict between memorization and instruction. The high probability the default value holds triggers the LLM to anchor to it, something that is not applicable in the FF case, in which the LLM has to generate a response without any reference to the original value in the prompt.

\paragraph{Assignment vs swapping}
There is a clear distinction between $R_a$ (\textit{assignment}) and $R_s$ (\textit{swapping}) cases: Swapping causes the LLMs to respond with the original constant value more frequently; we hypothesize that this occurs because the LLM’s memory is triggered, associating both constants with their default values and thus more readily ignoring  redefinition. Notably, this behavior remains consistent across all prompting methods tested.

\paragraph{The influence of prompting}

\begin{figure} [t!]
\vskip -0.08in
\centering 
\includegraphics[width=0.95\linewidth]{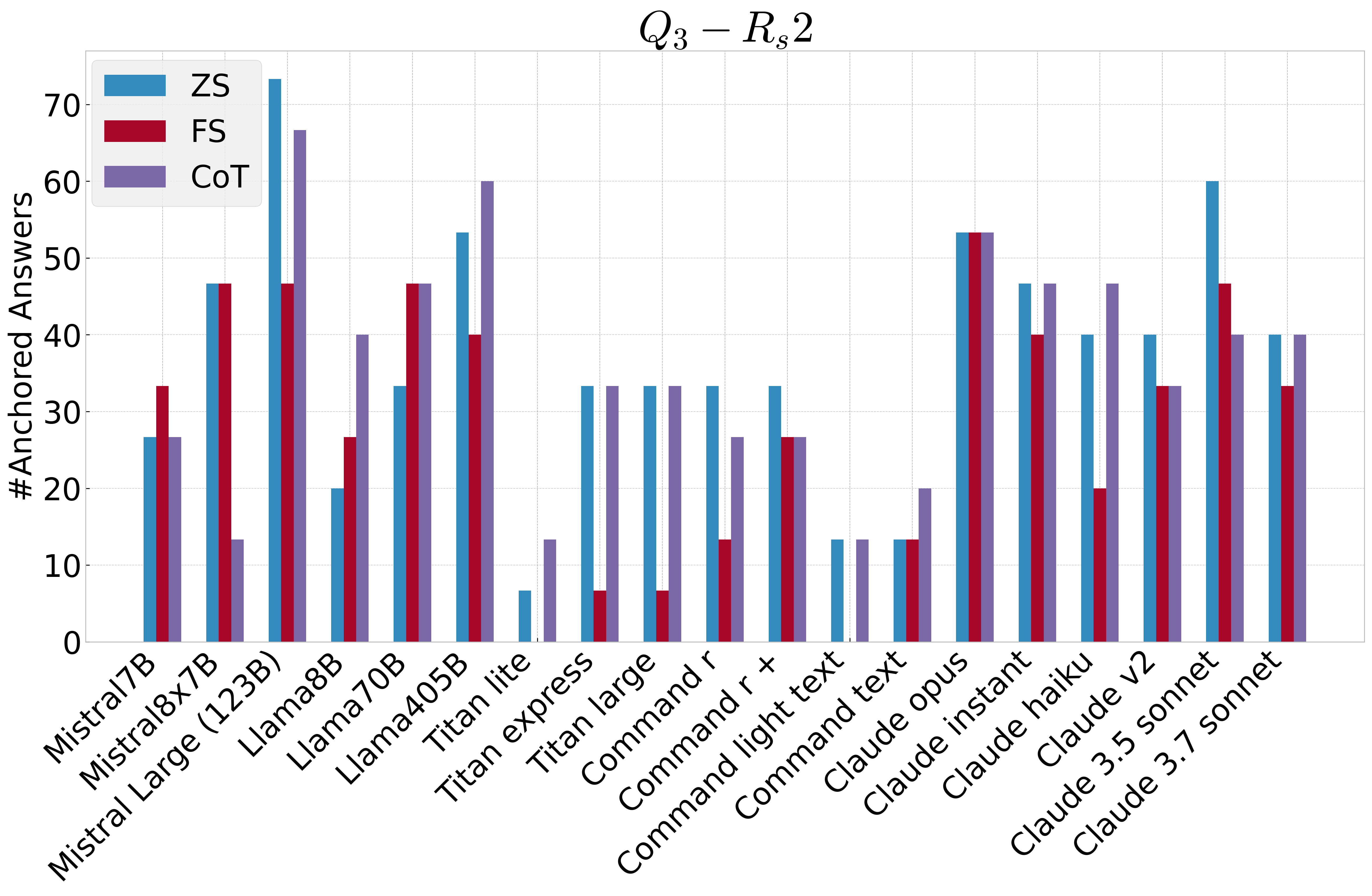} 
\vskip -0.07in
\caption{Comparison of the anchored response rate for $Q_3$ questions in the $R_{s2}$ redefinition level for all LLMs.} \label{fig:model_zs_fs_cot} 
\vskip -0.1in
\end{figure}

Figure \ref{fig:model_zs_fs_cot} highlights the role of prompting in driving the anchoring rate of different LLMs and prompting techniques on $Q_3$ questions and the hardest $R_s2$ redefinition level regarding  \textit{swapping}. Interestingly,  CoT prompting does not help LLMs (even larger ones) avoid anchoring or force them to follow the redefinition task. Instead, FS prompting proves to be more effective in most cases, with 50\% of the LLMs tested achieving the minimum anchoring rate using FS.
Certain LLMs, such as Mixtral 8x7B/Large, Titan Large, and Claude Haiku, exhibit a significant variance between the maximum and minimum number of anchored responses depending on the prompting technique used. However, this is not a consistent pattern, as most LLMs have a comparable anchoring rate across different techniques. Specifically,  the average difference between the maximum and minimum anchoring rate for all LLMs is 16.29 ± 9.22\%, indicating that \textit{prompting generally has a relatively small impact} on this phenomenon. Similar behaviors are observed for the other redefinition and question levels. 

\paragraph{Refusal to respond}
\label{sec:refusal}
In some cases,  a portion of the LLMs' completely wrong responses does not stem from reasoning inability, but rather from their refusal to perform the redefinition. By refusing to respond, the LLM showcases robustness, since it cannot be misled by a possibly malicious redefinition; however, this behavior obscures the LLM's actual reasoning abilities. To quantify this, we measure LLM refusal rates by  categorizing wrong responses into two groups: (i) \textbf{actually wrong} and (ii) \textbf{redefinition refusal} responses. Table \ref{tab:refusal_rate_main} presents the average refusal rate among wrong responses for all question levels. It is evident that \textit{LLM refusal rates vary significantly}, with LLama and Mistral emerging as the LLM families with higher refusal rates. Also, larger models refuse less often, indicating a \textit{false confidence towards providing a response}. Ultimately, refusal is primarily related to LLM scale rather than NR reasoning abilities, with correlations between  refusal rate and NR accuracy being weak (0.144 and 0.039 on average for the FF and MC response formats respectively).

\begin{table}[t!]
\centering
\small
\begin{tabular}{p{1.7cm}|p{0.75cm}|c|c}
\hline
Model                  & Prompt & FF & MC \\ \hline

      \multirow{3}{*}{Mistral7B}  &  ZS & \underline{6.57 ± 11.99} & 13.34 ± 18.07 \\
 &  CoT & 5.63 ± 8.89 & \underline{15.62 ± 16.45} \\
 &  FS & \textbf{3.7 ± 7.58} & \textbf{10.07 ± 15.25} \\
\hline
\multirow{3}{*}{Mixtral8x7B}  &  ZS & \underline{18.0 ± 22.8} & 8.61 ± 16.97 \\
 &  CoT & 9.22 ± 16.82 & \underline{15.5 ± 17.63} \\
 &  FS & \textbf{10.98 ± 17.03} & \textbf{5.95 ± 18.79} \\
\hline
\multirow{3}{*}{Mistral Large}  &  ZS & \underline{16.33 ± 33.69} & 1.67 ± 6.24 \\
 &  CoT & \textbf{8.33 ± 18.51} & \textbf{0 ± 0} \\
 &  FS & 14.35 ± 26.96 & \underline{1.33 ± 4.99} \\
\hline
\multirow{3}{*}{Llama8B}  &  ZS & \underline{55.54 ± 24.37} & \underline{40.05 ± 18.58} \\
 &  CoT & 35.25 ± 23.33 & 32.89 ± 23.21 \\
 &  FS & \textbf{2.41 ± 6.64} & \textbf{0 ± 0}\\
\hline
\multirow{3}{*}{Llama70B}  &  ZS & \underline{38.66 ± 29.92} & 5.56 ± 14.49 \\
 &  CoT & 9.17 ± 17.36 & \underline{13.33 ± 27.35} \\
 &  FS & \textbf{0 ± 0} & \textbf{0 ± 0}\\
\hline
\multirow{3}{*}{Llama405B}  &  ZS & \underline{1.33 ± 4.99}  & \textbf{0 ± 0}\\
 &  CoT & \textbf{0 ± 0} & \textbf{0 ± 0} \\
 &  FS & \textbf{0 ± 0} & \textbf{0 ± 0} \\
\hline
\multirow{3}{*}{Titan lite}  &  ZS & \textbf{1.56 ± 3.19} & \textbf{0 ± 0}\\
 &  CoT & \underline{3.03 ± 5.66} & \textbf{0 ± 0}\\
 &  FS & 2.54 ± 5.39 & \textbf{0 ± 0} \\
\hline
\multirow{3}{*}{Titan express}  &  ZS & 0.56 ± 2.08  & \textbf{0 ± 0} \\
 &  CoT & \underline{1.9 ± 7.13} & \textbf{0 ± 0} \\
 &  FS & \textbf{0 ± 0} & \textbf{0 ± 0} \\ \hline
\multirow{3}{*}{Titan large}  &  ZS & \underline{2.0 ± 5.42}  & \textbf{0 ± 0}\\
 &  CoT & \textbf{0 ± 0} & \textbf{0 ± 0}\\
 &  FS & \textbf{0 ± 0} & \textbf{0 ± 0}\\
\hline
\multirow{3}{*}{Command text}  &  ZS & \underline{3.33 ± 9.03}  & \textbf{0 ± 0} \\
 &  CoT & \textbf{0 ± 0} & \textbf{0 ± 0}\\
 &  FS & 0.83 ± 3.12  & \textbf{0 ± 0} \\
\hline
\multirow{3}{*}{Claude Instant}  &  ZS & \underline{1.69 ± 4.36}  & \textbf{0 ± 0} \\
 &  CoT & \textbf{0 ± 0} & \textbf{0 ± 0}\\
 &  FS & 4.07 ± 12.58  & \textbf{0 ± 0} \\
\hline
\multirow{3}{*}{\shortstack{Claude v2}}  &  ZS & \underline{20.48 ± 26.25} & 4.83 ± 9.29 \\
 &  CoT & 14.31 ± 24.39 & \underline{10.0 ± 27.08} \\
 &  FS & \textbf{8.91 ± 24.75} & \textbf{3.17 ± 8.81} \\
\hline
\end{tabular}
\caption{Average refusal rates over all question levels (lowest values in \textbf{bold} and highest values \underline{underlined}). We exclude LLMs with zero refusal rate overall.}
\label{tab:refusal_rate_main}
\vskip -0.1in
\end{table}

Furthermore,  prompting techniques play a crucial role in refusal rates, with FS mitigating refusal the most. This result is intuitive, as the LLM is exposed to more examples containing redefinitions in its input, making it less likely to refuse the task. Additional results are provided in Appendix \ref{app:refusal}.

\begin{figure*}[t!]
\begin{subfigure}{\textwidth}
        \centering
         \includegraphics[width=0.86\linewidth]{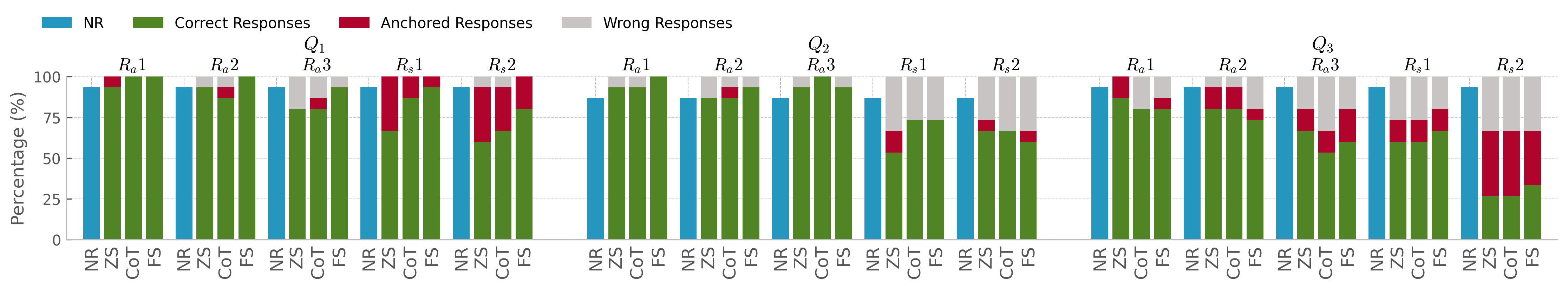}
         \vskip -0.01in
        \caption{Claude 3.7 Sonnet without Thinking in the FF response format.}
        \label{fig:no_thinking_constants_FF}
    \end{subfigure}
    
    \begin{subfigure}{\textwidth}
        \centering
        \includegraphics[width=0.86\linewidth]{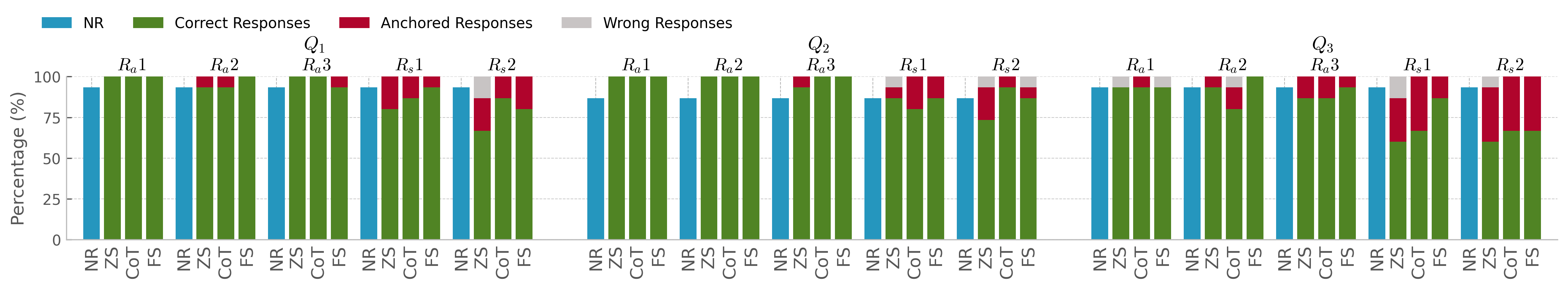}
        \vskip -0.01in
        \caption{Claude 3.7 Sonnet without Thinking in the MC response format.}
        \label{fig:no_thinking_constants_MC}
    \end{subfigure}
        \begin{subfigure}{\textwidth}
        \centering
        \includegraphics[width=0.86\linewidth]{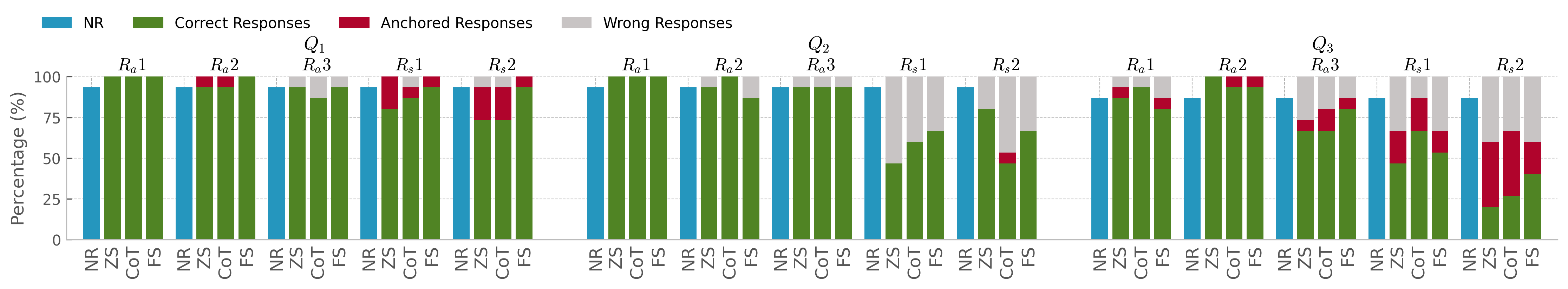}
        \vskip -0.01in
        \caption{Claude 3.7 Sonnet with Thinking in the FF response format.}
        \label{fig:thinking_constants_FF}
    \end{subfigure}

     \begin{subfigure}{\textwidth}
        \centering
        \includegraphics[width=0.86\linewidth]{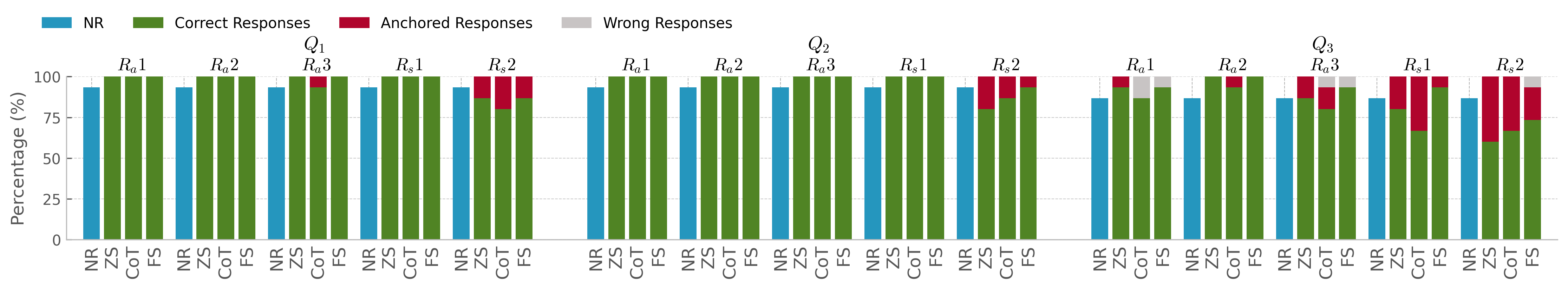}
        \vskip -0.01in
        \caption{Claude 3.7 Sonnet with Thinking in the MC response format.}
        \label{fig:thinking_constants_MC}
        \end{subfigure}
    \caption{Claude 3.7 Sonnet results without and with Thinking.}
    \label{fig:thinking}
\end{figure*}
\paragraph{The impact of thinking}
In the emergence of extended thinking modes in SoTA LLMs, such as Claude 3.7 Sonnet, we are able to elicit deeper reasoning processes of the model at hand, enabling better responses in multi-step reasoning situations, as in the case of elevated redefinition levels and question difficulty. Related findings are presented in Figure \ref{fig:thinking}. As observed, \textit{the contribution of thinking is minimal}, only slightly reducing the anchoring response rate in specific cases. Therefore, we can conclude that even advanced reasoning mechanisms cannot assist Claude 3.7 Sonnet in following the redefined thought processes needed to accurately respond, further strengthening the claim that the redefinition task challenges inherent reasoning functions and limitations of SoTA LLMs.
\subsection{Results on units of measure redefinition}
The findings on the redefinition of units of measure align with those of constants, also revealing a strong inverse trend: larger models (e.g., Mistral Large, Llama405B) consistently exhibit a higher anchoring rate compared to their smaller counterparts (Mistral7B, Llama8B), regardless of the response format,  redefinition level, or prompting method. This trend is particularly notable in $Q_3$ questions. However, anchoring is relatively lower than in the constants case, likely because the actual values of units are not commonly used in calculations, making the LLMs less prone to anchoring. A thorough analysis is presented in Appendix \ref{app:units_redefinition}.

\section{Conclusion}
\vskip -0.01in
In this work, we thoroughly investigate the redefinition task by prompting LLMs to reason with redefined values of physical constants and units of measure. We uncover several critical patterns in LLMs, showcasing pitfalls of scale, such as decreased reasoning capacity and increased confidence in erroneous answers instead of abstaining. Moreover, we offer extensive insights into how redefinition difficulty, prompting strategies, and response format influence LLMs’ propensity to anchor on their prior knowledge rather than reason flexibly.

\section*{Limitations}
Our redefinitions focus on concept sets with a well-defined number of elements, such as mathematical constants and unit measures, restricting our investigation to closed-world reasoning rather than broader, more generalizable tasks. We opt for such restricted settings in order to evaluate more clearly the impact of redefinition difficulty in conjunction to question difficulty, targeting certain LLM reasoning capabilities. Evaluating additional types of reasoning over redefinitions should consider more broad concept sets to be redefined.

Furthermore, we do not compare LLM performance on redefinition tasks with human performance, following the assertion of \citet{mckenzie2024inversescalingbiggerisnt} that such tasks are generally easy for humans, albeit requiring some effort in complex cases.
Conducting a human experiment for direct comparison would be highly challenging due to significant variability in individual knowledge and expertise. Prior exposure to related tasks could further bias results—for example, a physics teacher may solve redefinition problems with ease, whereas others may struggle. Moreover,  concentration, memory, motivation, engagement, psychological and environmental factors play a decisive role in human performance, making controlled experimentation significantly difficult and possibly unreliable.

In terms of prompting, more refined strategies can be tested, especially in the FS case, where sophisticated prompting techniques have been recently proposed to enhance exemplar selection quality via reasoning similarity \cite{panagiotopoulos-etal-2025-riscore} in place of the generic semantic similarity measure. 

\section*{Ethical considerations}
The ability to redefine concepts in LLMs presents ethical challenges, particularly in the generation of misleading or deceptive responses. Our study highlights an inherent trade-off between reasoning transparency and model robustness. More robust models resist redefinition by refusing the task, making them less susceptible to manipulation but also limiting their ability to engage in flexible reasoning. Conversely, models that effectively reason with redefined values exhibit greater transparency and adaptability but are more vulnerable to malicious prompts. This duality raises a significant ethical question: should LLMs prioritize strict factual adherence at the cost of reasoning flexibility, or should they remain adaptable at the risk of being misled? Addressing this trade-off is a crucial ethical consideration in the responsible design and deployment of LLMs in general.


\bibliography{main}
\appendix
\section{$Q_1$ exceptions}
\label{sec:exceptions-q1}
Regarding the constants redefinition task, we consider some exceptions to the default $Q_1$: \textit{What is the first -non-zero- digit of \{constant\}?}. These exceptions are listed in Table \ref{tab:exceptions-q1}. Those questions directly trigger the existing knowledge of LLMs, mostly requesting retrieving rather than reasoning on the queried information. For example, it is well known that $i^2=-1$ and there is no reasoning on the question \textit{$Q_1$: What is the value of $i^2$?}.

\begin{table}[h!]
    \centering \small
    \begin{tabular}{c|c}
\hline
    & $Q_1$ \\
\hline
$c$&  How far does light travel in one second? \\
$i$ & What is the value of $i^2$? \\
$\infty$ & What is the value of infinity? \\
zero & What is the absolute value of zero? \\
\hline
    \end{tabular}
    \caption{Exceptions to the typical $Q_1$ format}
    \label{tab:exceptions-q1}
\end{table}
\section{Model Knowledgeability}
\label{app:correlation}

Tables \ref{tab:correlation_few} and \ref{tab:correlation_cot} show the correlation between performance before redefinition (NR case) and the anchored response rate for each constant  redefinition and question level in the FS and CoT prompting setups respectively. The pattern remains the same: in $Q_1$ and $Q_2$ question levels, there is little to no correlation or a negative correlation between the two values, whereas in $Q_3$—particularly in the \textit{swapping} case—the correlation is highly positive. This suggests that in those cases, more knowledgeable models adapt less to the redefinition.

Table \ref{tab:anchored_table_NR_all} presents the number of correct answers in the NR case alongside the percentage of anchored responses for constant redefinitions across all LLMs used in the study.

\begin{table}[h]
\small
    \centering
    \begin{tabular}{l|ccc|cc}
        \hline
        Level & $R_a1$ & $R_a2$ & $R_a3$ & $R_s1$ & $R_s2$ \\ \hline
        \multicolumn{6}{c}{Free-Form (FF)} \\ 
         \hline
        $Q_1$ & -0.055 & -0.129 & \cellcolor{lightdustygreen} -0.472 & 0.235 & -0.008 \\ 
        $Q_2$ & -0.283 & \cellcolor{lightdustygreen} -0.359 &  \cellcolor{lightdustygreen} -0.444 & 0.085 & -0.148 \\
        $Q_3$ & \cellcolor{lightdustypink} 0.356 & \cellcolor{lightdustypink} 0.374 & \cellcolor{lightdustypink} 0.492 & \cellcolor{lightdustypink} 0.596 & \cellcolor{lightdustypink} 0.823 \\
        \hline
        \multicolumn{6}{c}{Multiple Choice (MC)} \\ \hline
        $Q_1$ & \cellcolor{lightdustygreen} -0.71 & \cellcolor{lightdustygreen} -0.624 & \cellcolor{lightdustygreen} -0.711 & \cellcolor{lightdustygreen} -0.304 &  -0.28 \\
        $Q_2$ & -0.258 & \cellcolor{lightdustygreen} -0.473 & \cellcolor{lightdustygreen} -0.312 & \cellcolor{lightdustypink} 0.441 & -0.15 \\ 
        $Q_3$ & 0.269 & \cellcolor{lightdustypink} 0.589 & 0.288 & \cellcolor{lightdustypink} 0.624 & \cellcolor{lightdustypink} 0.694 \\
        \hline
    \end{tabular}
    \caption{Correlation between model performance before redefinition with the percentage of anchored answers for each type of constant redefinition and question level in FS setup. 
    Cells highlighted in \textcolor{lightdustypink}{pink} indicate a \textbf{high positive correlation} ($>0.3$), while cells in \textcolor{lightdustygreen}{green} indicate a \textbf{high negative correlation} ($<-0.3$).}
    \label{tab:correlation_few}
\end{table}
\begin{table}[h]
\small
    \centering
    \begin{tabular}{l|ccc|cc}
        \hline
        Level & $R_a1$ & $R_a2$ & $R_a3$ & $R_s1$ & $R_s2$ \\ \hline
        \multicolumn{6}{c}{Free-Form (FF)} \\ 
        \hline
        $Q_1$ & \cellcolor{lightdustygreen} -0.539 & \cellcolor{lightdustygreen} -0.542 & \cellcolor{lightdustygreen} -0.552 & -0.244 & \cellcolor{lightdustygreen} -0.319 \\
        $Q_2$ & \cellcolor{lightdustygreen} -0.521 & \cellcolor{lightdustygreen} -0.626 & \cellcolor{lightdustygreen} -0.58 & 0.143 & -0.125 \\ 
        $Q_3$ & \cellcolor{lightdustypink} 0.41 & 0.116 & -0.085 &  \cellcolor{lightdustypink} 0.71 &  \cellcolor{lightdustypink} 0.588 \\ 
        \hline
        \multicolumn{6}{c}{Multiple Choice (MC)} \\ \hline
$Q_1$ & \cellcolor{lightdustygreen} -0.529 & \cellcolor{lightdustygreen} -0.483 & \cellcolor{lightdustygreen} -0.358 & -0.17 & 0.16 \\
$Q_2$  & -0.183 & -0.224 & -0.202 & \cellcolor{lightdustypink} 0.329 & -0.044 \\
 $Q_3$ & 0.134 & \cellcolor{lightdustypink} 0.366 & 0.009 & \cellcolor{lightdustypink} 0.679 & \cellcolor{lightdustypink} 0.657 \\
        \hline
    \end{tabular}
    \caption{Correlation between model performance before redefinition with the percentage of anchored answers for each type of constant redefinition and question level in CoT setup. 
    Cells highlighted in \textcolor{lightdustypink}{pink} indicate a \textbf{high positive correlation} ($>0.3$), while cells in \textcolor{lightdustygreen}{green} indicate a \textbf{high negative correlation} ($<-0.3$).}
    \label{tab:correlation_cot}
\end{table}

\begin{table*}[h]
\centering
\small
\begin{tabular}{l|p{0.7cm}p{0.6cm}|p{0.7cm}p{0.6cm}|p{0.7cm}p{0.6cm}|p{0.7cm}p{0.6cm}|p{0.7cm}p{0.6cm}|p{0.7cm}p{0.6cm}}
\hline
\multirow{3}{*}{Model} & \multicolumn{6}{c|}{$R_a3$}                                                               & \multicolumn{6}{c}{$R_s2$}                                                                \\ \cline{2-13}
                       & \multicolumn{2}{c}{$Q_1$} & \multicolumn{2}{c}{$Q_2$} & \multicolumn{2}{c|}{$Q_3$} & \multicolumn{2}{c}{$Q_1$} & \multicolumn{2}{c}{$Q_2$} & \multicolumn{2}{c}{$Q_3$} \\ \cline{2-13}
                       & NR           & FF           & NR           & FF           & NR            & FF           & NR           & FF           & NR           & FF           & NR           & FF           \\ \hline

Mistral7B & 66.67 & 33.33 & 46.67 & 33.33 & 33.33 & 26.67 & 66.67 & 33.33 & 46.67 & 13.33 & 33.33 & 26.67 \\ 
Mixtral8x7B & 100.0 & 33.33 & 66.67 & 26.67 & 66.67 & 20.0 & 100.0 & 26.67 & 66.67 & 40.0 & 66.67 & 46.67 \\ 
Mistral Large (123B) & 93.33 & 33.33 & 73.33 & 26.67 & 53.33 & 53.33 & 93.33 & 66.67 & 73.33 & 46.67 & 53.33 & 73.33 \\  \hline
Llama8B & 80.0 & 0.0 & 80.0 & 0.0 &  53.33 & 13.33 & 80.0 & 20.0 & 80.0 & 26.67 & 53.33 & 20.0 \\ 
Llama70B & 93.33 & 6.67 & 80.0 & 0.0 & 80.0 & 13.33 & 93.33 & 33.33 & 80.0 & 13.33 & 80.0 & 33.33 \\ 
Llama405B & 93.33 & 0.0 & 86.67 & 0.0 & 73.33 & 26.67 & 93.33 & 26.67 & 86.67 & 6.67 & 73.33 & 53.33 \\  \hline
Titan lite & 46.67 & 13.33 & 20.0 & 20.0 & 20.0 & 0.0 & 46.67 & 40.0 & 20.0 & 20.0 & 20.0 & 6.67 \\
Titan express & 73.33 & 20.0 & 33.33 & 13.33 & 26.67 & 20.0 & 73.33 & 40.0 & 33.33 & 20.0 & 26.67 & 33.33 \\
Titan large & 66.67 & 26.67 & 33.33 & 20.0 & 26.67 & 13.33 & 66.67 & 60.0 & 33.33 & 13.33 & 26.67 & 33.33 \\  \hline
Command r & 86.67 & 0.0 & 33.33 & 20.0 & 40.0 & 26.67 & 86.67 & 53.33 & 33.33 & 20.0 & 40.0 & 33.33 \\ 
Command r + & 93.33 & 6.67 & 66.67 & 0.0 & 66.67 & 13.33 & 93.33 & 13.33 & 66.67 & 26.67 & 66.67 & 33.33 \\ 
Command light text & 60.0 & 6.67 & 6.67 & 13.33 & 0.0 & 0.0 & 60.0 & 13.33 & 6.67 & 26.67 & 0.0 & 13.33 \\ 
Command text & 53.33 & 13.33 & 40.0 & 6.67 & 26.67 & 6.67 & 53.33 & 40.0 & 40.0 & 13.33 & 26.67 & 13.33 \\ \hline
Claude Opus & 100.0 & 13.33 & 80.0 & 6.67 & 80.0 & 33.33 & 100.0 & 46.67 & 80.0 & 20.0 & 80.0 & 53.33 \\  
Claude Instant & 86.67 & 0.0 & 33.33 & 13.33 & 46.67 & 26.67 & 86.67 & 33.33 & 33.33 & 33.33 & 46.67 & 46.67 \\ 
Claude Haiku & 100.0 & 20.0 & 73.33 & 6.67 & 66.67 & 20.0 & 100.0 & 26.67 & 73.33 & 20.0 & 66.67 & 40.0 \\ 
Claude v2 & 73.33 & 26.67 & 60.0 & 20.0 & 40.0 & 46.67 & 73.33 & 13.33 & 60.0 & 33.33 & 40.0 & 40.0 \\
Claude 3.5 Sonnet & 100.0 & 26.67 & 93.33 & 0.0 & 86.67 & 13.33 & 100.0 & 33.33 & 93.33 & 20.0 & 86.67 & 60.0 \\ 
Claude 3.7 Sonnet & 93.33 & 0.0 & 86.67 & 0.0 & 93.33 & 13.33 & 93.33 & 33.33 & 86.67 & 6.67 & 93.33 & 40.0 \\ 
\hline
\end{tabular}

\caption{The percentage of correct responses with no redefinition (NR) and the anchored responses after constant redefinitions regarding free-form (FF) responses and using ZS prompting.}
\label{tab:anchored_table_NR_all}
\end{table*}

\section{Inverse Trends}
\label{app:inverse}

\begin{figure} [h!]
\centering 
\includegraphics[width=\linewidth]{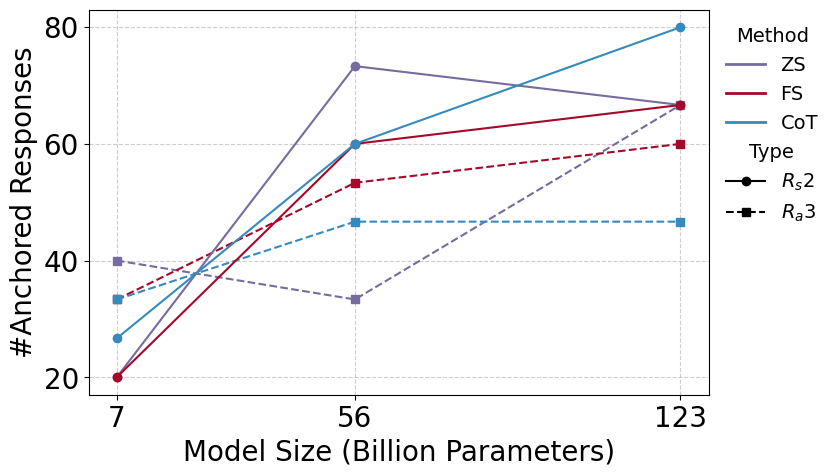} 
\caption{Anchored response rate after constants redefinitions for LLMs of varying sizes in the Mistral family under the MC response format, harnessing different prompting techniques on the hardest redefinition levels.} \label{fig:model_size_vs_anchored_responses_mistral} 
\end{figure}

\begin{figure*}[h!]
\begin{subfigure}{\textwidth}
        \centering
         \includegraphics[width=0.86\linewidth]{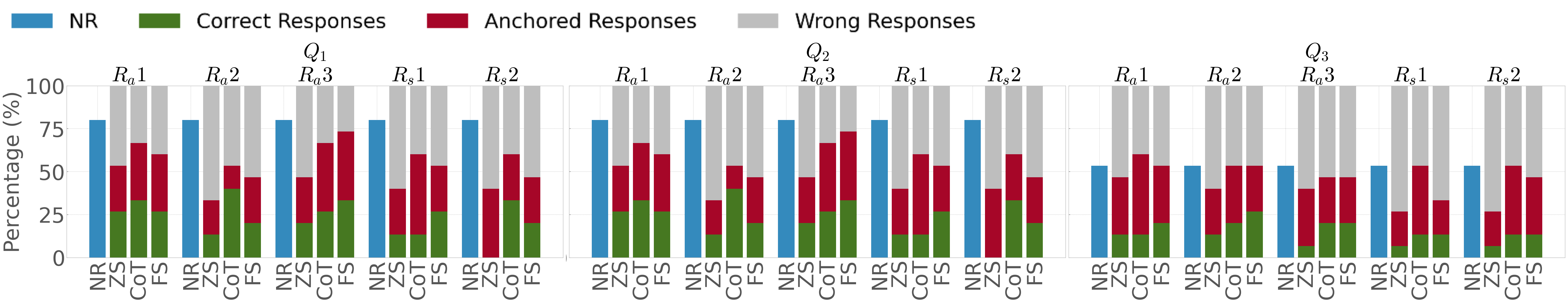}
         \vskip -0.01in
        \caption{Response breakdown for Llama 8B before and after constant redefinitions.}
        \label{fig:llama8b_MC}
    \end{subfigure}
    
    \begin{subfigure}{\textwidth}
        \centering
        \includegraphics[width=0.86\linewidth]{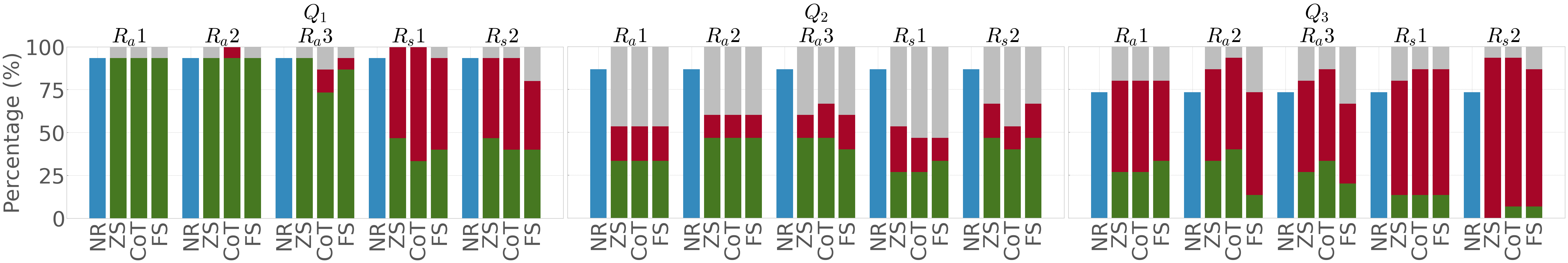}
        \vskip -0.01in
        \caption{Response breakdown for Llama 405B before and after constant redefinitions.}
        \label{fig:mistral_large_MC}
    \end{subfigure}
    \vskip -0.01in
    \caption{Comparison of Llama 8B and Llama 405B  responses on the MC response format.}
    \label{fig:llama_all}
\end{figure*}

Figure \ref{fig:llama_all} shows the rate of correct answers in the NR task, as well as the rates of correct, wrong, and anchored responses for Llama 8B and Llama 405B after the constant redefinition task in the MC response format. From this figure, it is observable that the same pattern as in Mistral (Figure \ref{fig:mistral_all}) emerges, where the number of anchored responses is significantly higher in the larger model.  
Additionally, Figure \ref{fig:model_size_vs_anchored_responses_mistral} shows the number of anchored responses after constants redefinitions for models of varying sizes in the Mistral family in the MC response format. The trend is the same as in Llama models illustrated in the main paper (Figure \ref{fig:model_size_vs_anchored_responses}), where larger variants tend to anchor more to their prior knowledge in comparison to the smaller ones, for both $R_s2, R_a3$ redefinition levels and all prompting techniques. Mixtral7x8B in the ZS prompting setup produces only slightly more anchored responses compared to the larger variant (Mistral Large (123B parameters)). However, this is likely due to the increased performance of the LLM in the NR case, as shown in Table \ref{tab:anchored_table_NR}. The difference is relatively small, meaning that while there is a measurable increase in anchoring for Mixtral7x8B, it is not substantial enough to indicate a significant deviation from the expected trend. 

\section{Refusal Rates}
\label{app:refusal}
An important observation stated in the main paper is that completely wrong responses  include instances where the LLM actively refuses to respond to the redefined problem.

\begin{figure*}[h!]
\begin{subfigure}{\textwidth}
        \centering
         \includegraphics[width=0.86\linewidth]{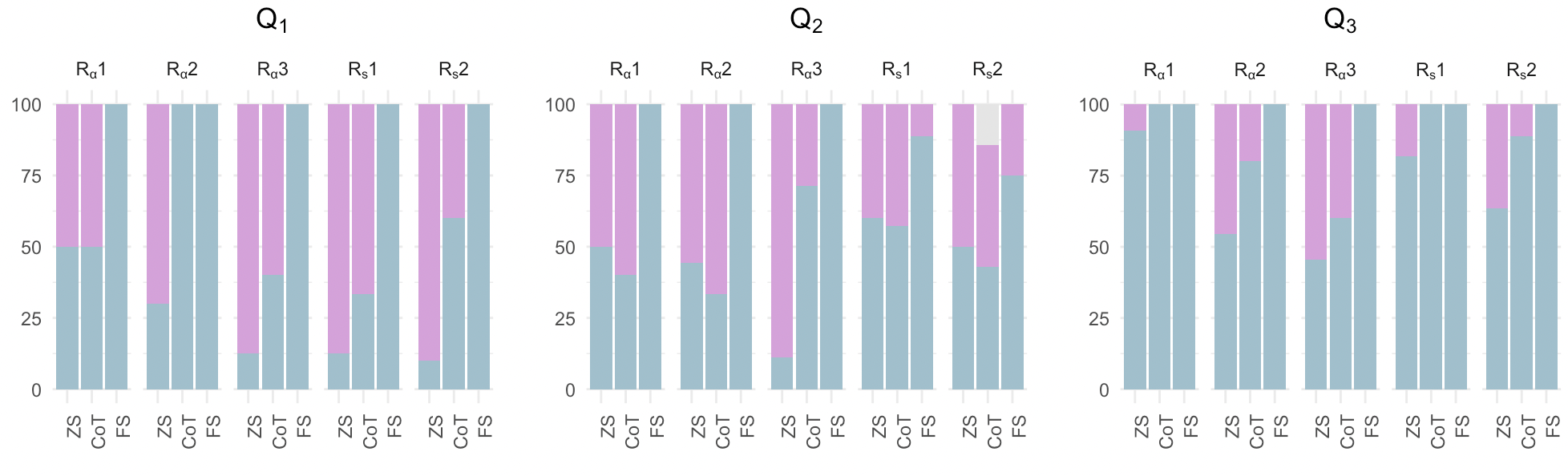}
         \vskip -0.01in
    \caption{Response breakdown for Llama 8B FF responses.}
    \end{subfigure}
    
    \begin{subfigure}{\textwidth}
        \centering
        \includegraphics[width=0.86\linewidth]{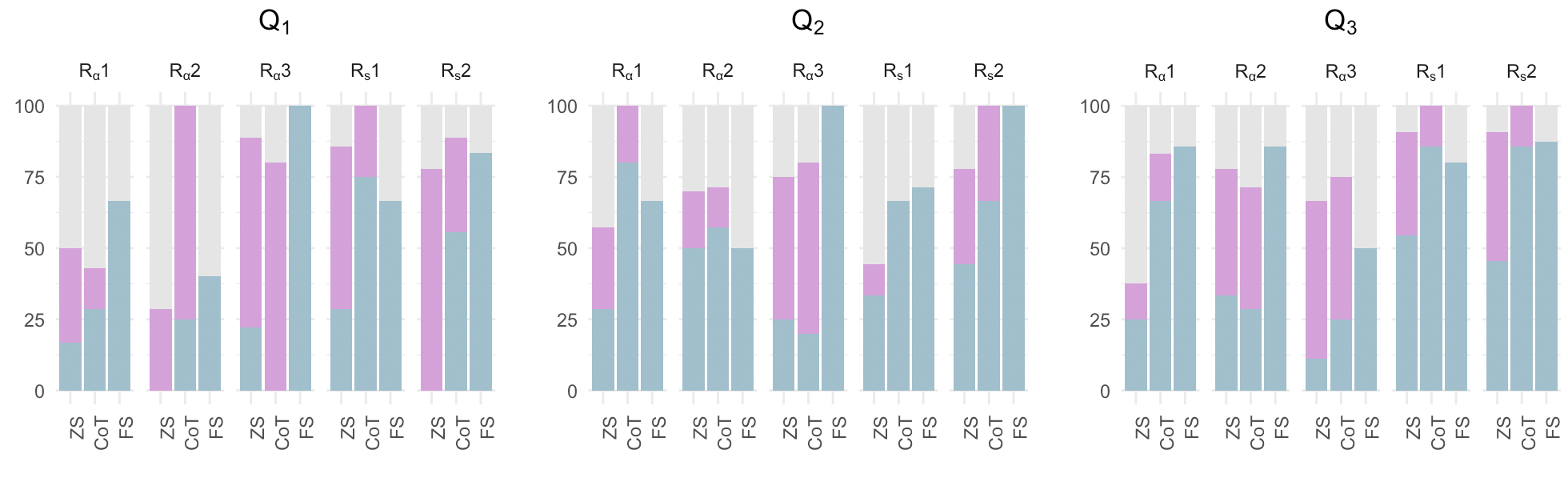}
        \vskip -0.01in
        \caption{Response breakdown for Llama 8B MC responses.}
    \end{subfigure}
    \vskip -0.01in
    \caption{Completely wrong responses breakdown for Llama 8B. 
\textcolor{dustyblue}{Blue} denotes actually wrong responses,
\textcolor{dustypurple}{Purple} indicates refusals, while \textcolor{dustygray}{Gray} instances correspond to blank responses}.
    \label{fig:llama8b-mc-refusal}
\end{figure*}

\begin{figure*}[h!]
\begin{subfigure}{\textwidth}
        \centering
         \includegraphics[width=0.86\linewidth]{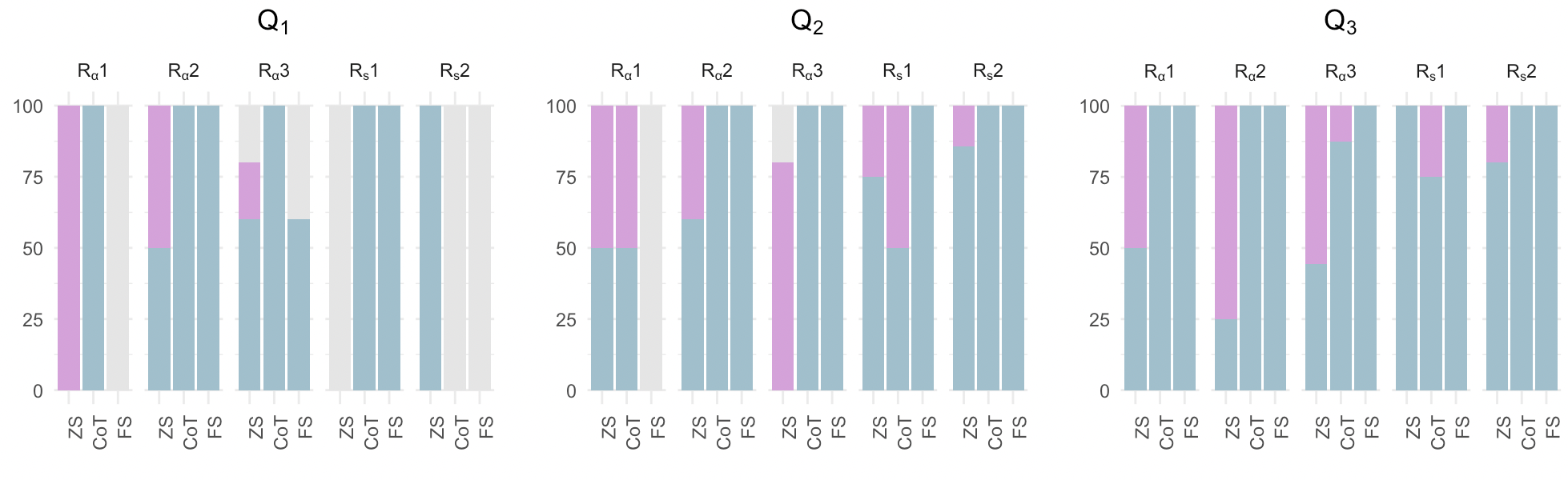}
         \vskip -0.01in
    \caption{Response breakdown for Llama 70B FF responses.}
    \end{subfigure}
    
    \begin{subfigure}{\textwidth}
        \centering
        \includegraphics[width=0.86\linewidth]{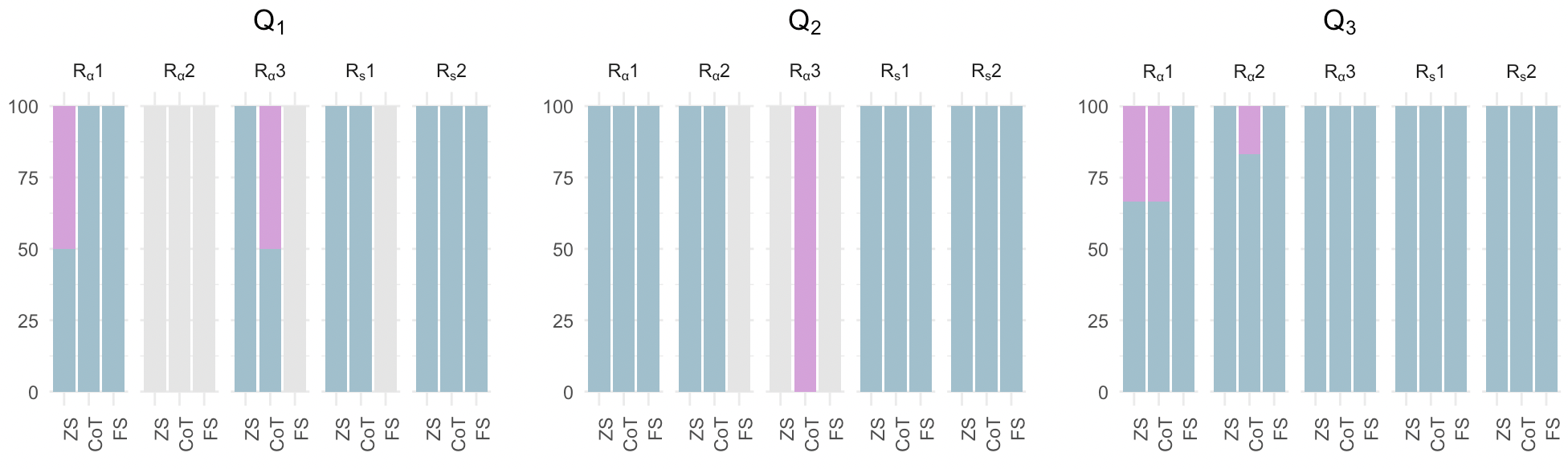}
        \vskip -0.01in
        \caption{Response breakdown for Llama 70B MC responses.}
    \end{subfigure}
    \vskip -0.01in
    \caption{Completely wrong responses breakdown for Llama 70B. 
\textcolor{dustyblue}{Blue} denotes actually wrong responses,
\textcolor{dustypurple}{Purple} indicates refusals, while \textcolor{dustygray}{Gray} instances correspond to blank responses.}.
    \label{fig:llama70b-mc-refusal}
\end{figure*}

\begin{figure*}[h!]
\begin{subfigure}{\textwidth}
        \centering
         \includegraphics[width=0.86\linewidth]{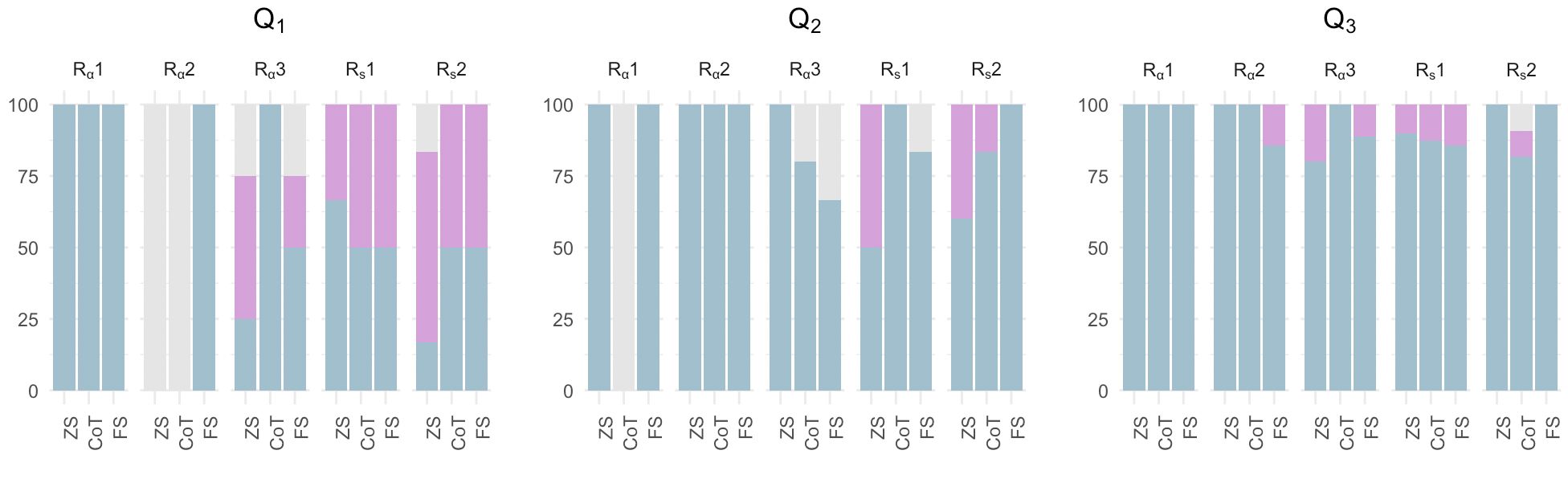}
         \vskip -0.01in
    \caption{Response breakdown for Mixtral8x7 FF responses.}
    \end{subfigure}
    
    \begin{subfigure}{\textwidth}
        \centering
        \includegraphics[width=0.86\linewidth]{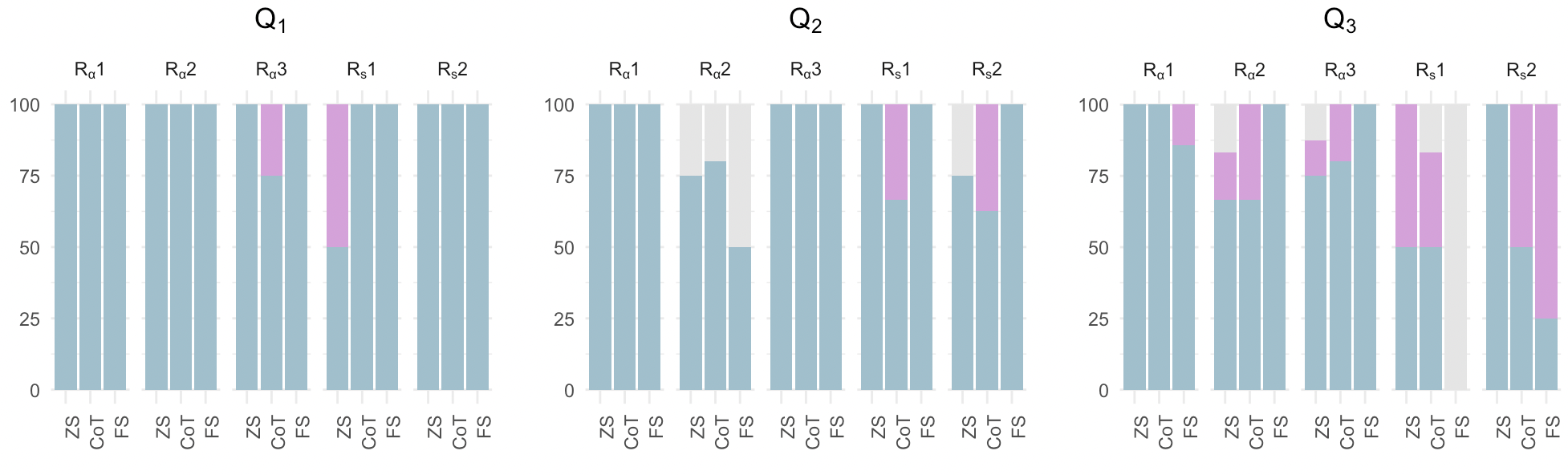}
        \vskip -0.01in
        \caption{Response breakdown for Mixtral8x7 MC responses.}
    \end{subfigure}
    \vskip -0.01in
    \caption{Completely wrong responses breakdown for Mixtral8x7. 
\textcolor{dustyblue}{Blue} denotes actually wrong responses,
\textcolor{dustypurple}{Purple} indicates refusals, while \textcolor{dustygray}{Gray} instances correspond to blank responses.}.
    \label{fig:mixtral-mc-refusal}
\end{figure*}

\begin{figure*}[h!]
\begin{subfigure}{\textwidth}
        \centering
         \includegraphics[width=0.86\linewidth]{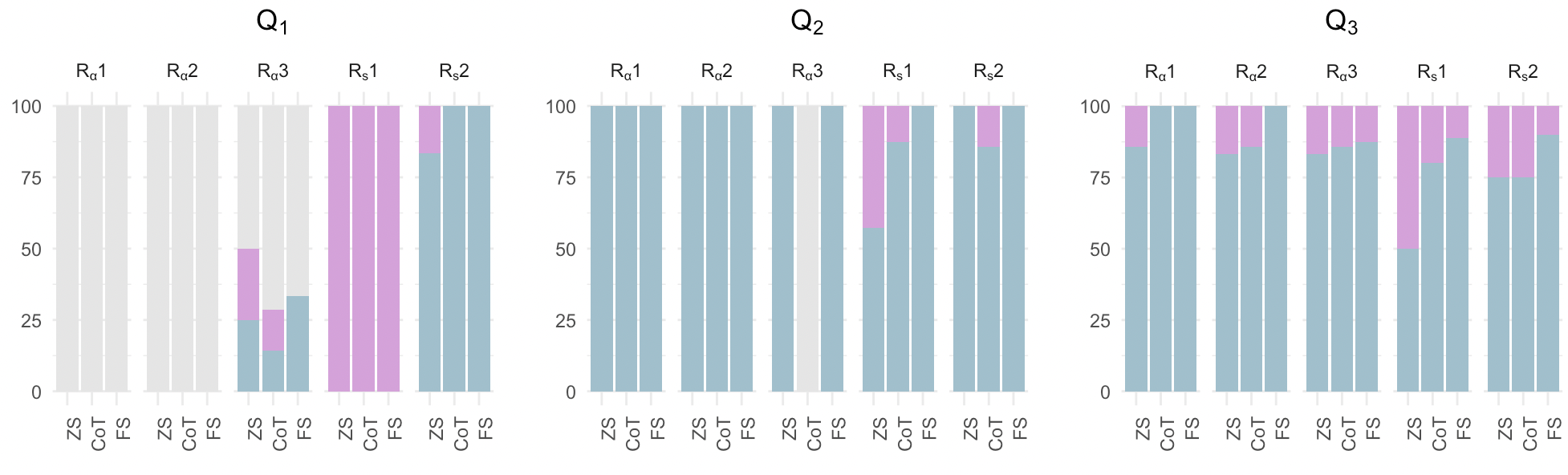}
         \vskip -0.01in
    \caption{Response breakdown for Claude v2 FF responses.}
    \end{subfigure}
    
    \begin{subfigure}{\textwidth}
        \centering
        \includegraphics[width=0.86\linewidth]{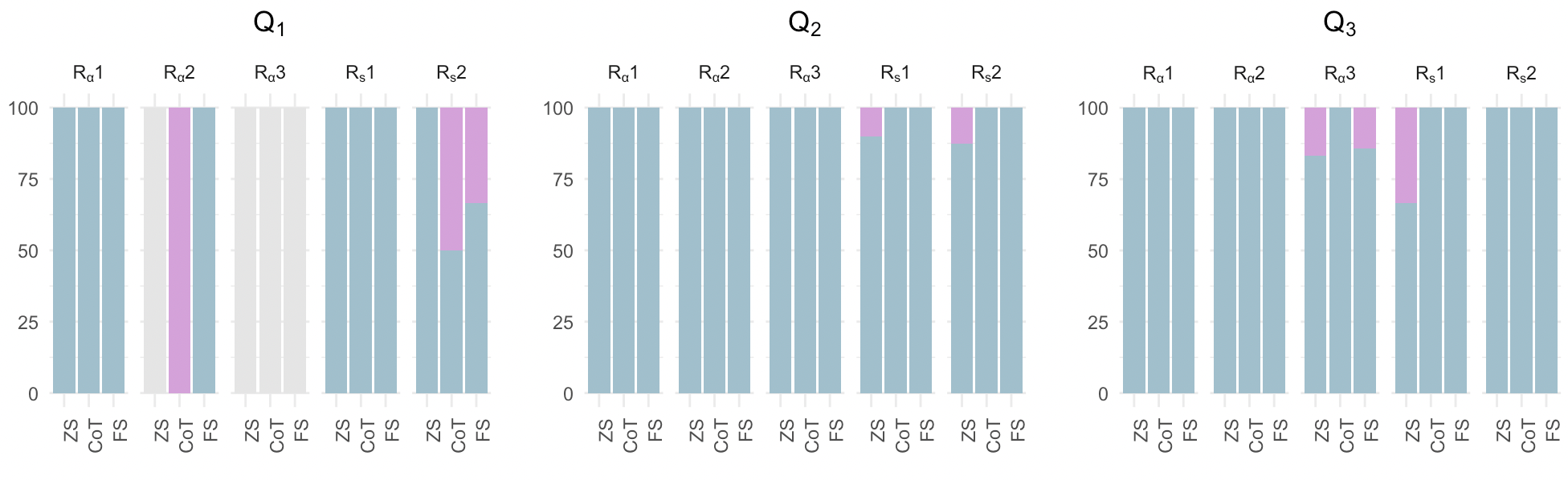}
        \vskip -0.01in
        \caption{Response breakdown for Claude v2 MC responses.}
    \end{subfigure}
    \vskip -0.01in
    \caption{Completely wrong responses breakdown for Claude v2. 
\textcolor{dustyblue}{Blue} denotes actually wrong responses,
\textcolor{dustypurple}{Purple} indicates refusals, while \textcolor{dustygray}{Gray} instances correspond to blank responses.}.
    \label{fig:claude-mc-refusal}
\end{figure*}

An analysis of the most interesting cases regarding wrong responses is presented in Figures \ref{fig:llama8b-mc-refusal}, \ref{fig:llama70b-mc-refusal}, \ref{fig:mixtral-mc-refusal}, \ref{fig:claude-mc-refusal} for selected LLMs. For example, in the case of Llama 8B (Figure \ref{fig:llama8b-mc-refusal}) the refusal rate diminishes as questions get harder. This reveals an overconfidence in responding instead of abstaining, leading to a problematic behavior, since Llama 8B achieves very few correct responses in all question levels, and especially in harder ones ($Q_3$), as indicated in Figure \ref{fig:llama8b_MC}. The difference between FF and MC response formats lies in the elevated number of blank responses in the MC case. This is because Llama 8B suffers when prompted to select one of the predefined options, indicating high uncertainty. Nevertheless, when prompted to respond without restrictions, empty responses are almost non-existent, revealing another sense of overconfidence tied to response format.

Conversely, Llama 70B (Figure \ref{fig:llama70b-mc-refusal}) refrains from empty responses, especially in harder cases, revealing its lower uncertainty in generating a response. Contrary to its smaller counterpart, its refusal rates decrease as questions get harder, leading to an increased number of actually wrong responses (reaching up to 100\% in some cases) over refusals.

A mixed behavior is presented in the case of Mixtral8x7B (Figure \ref{fig:mixtral-mc-refusal}), where refusal decreases in the FF response format, while it increases in the MC format. This behavior denotes that when Mixtral8x7B is exposed to a limited set of options, it elevates its resistance in redefining constants, possibly detecting the presented conflict between memorization and instruction. On the other hand, when generation is unrestricted, as in the FF case, its denial becomes significantly diminished, resulting in erroneous responses more often than not. Ultimately, in this case, response format is of outmost importance in defining the trade-off between response refusal and erroneous generations.

Finally, mixed patterns also occur in the case of Claude v2 (Figure \ref{fig:claude-mc-refusal}), where alternating patterns between 100\% refusal or 100\% actually wrong responses are revealed (as in the $Q_1$ question level). Therefore, this model is rather unpredictable in whether it prefers to deny the task or respond erroneously, since there is no obvious reason behind this diverging behavior. Contrary to the aforementioned Mixtral8x7 case, Claude v2 presents more wrong responses in the MC case in comparison to FF responses. That means that for some unexpected reason, it refuses to answer within an unrestricted setting, but results in wrong responses when presented with a limited number of options. This behavior reveals that Claude v2 confidently performs redefinitions with ease and without much resistance, but fails to solve the redefined problem overall.

Other than that, Tables \ref{tab:refusal-constants-q1}, \ref{tab:refusal-constants-q2}, and \ref{tab:refusal-constants-q3} present the proportion of incorrect responses attributed to the LLM's refusal to respond to the constants redefinition task over all completely wrong responses for all LLMs together. These Tables report refusal rates for each LLM, prompting method, and redefinition level regarding redefinitions across all $Q_1$, $Q_2$, and $Q_3$ question levels respectively. 

The results indicate that models such as Command r+, Command light, and Claude Haiku consistently tend to provide responses, ignoring their validity, therefore exhibiting \textit{no refusal instances}. This overconfidence is problematic in practice, even though useful in our experimentation, since reasoning shortcomings are exposed.
In contrast, models such as Llama, Mistral, and Claude v2 occasionally decline to generate a response when faced with the redefined task, possibly acknowledging their intrinsic inability to answer properly. This observation showcases that those LLMs that prefer to abstain from responding are more robust, since a redefinition prompt could act as a malicious adversarial attack that aims to mislead the LLM towards generating invalid responses. On the other hand, this deliberate action prevents them to reveal their reasoning capabilities, once again verifying the trade-off between robustness and evident reasoning.

Additionally, the type of prompting significantly influences the LLM's refusal rate. Specifically, models exhibit lower refusal rates in the FS setup compared to the ZS and CoT configurations. We assume that this is because LLMs may 'feel' overconfident when exposed to FS exemplars that clearly showcase the redefinition task to be performed, overriding their inherent inability to actually and properly reason over redefined concepts.

\begin{table*}[h!]
\small
\centering
\begin{tabular}{l|c|ccccc|ccccc}
\hline
\multirow{2}{*}{Model} & \multirow{2}{*}{Prompt} & \multicolumn{5}{c|}{FF}               & \multicolumn{5}{c}{MC}                \\
\cline{3-12}
                       &                         & $R_a1$ & $R_a2$ & $R_a3$ & $R_s1$ & $R_s2$ & $R_a1$ & $R_a2$ & $R_a3$ & $R_s1$ & $R_s2$ \\ \hline

\multirow{3}{*}{Mistral7B}  &  ZS  &  0.0  &  0.0  &  0.0  &  16.67  &  44.44  &  40.0  &  0.0  &  50.0  &  50.0  &  0.0 \\
 &  CoT  &  0.0  &  0.0  &  0.0  &  25.0  &  11.11  &  0.0  &  40.0  &  0.0  &  50.0  &  0.0 \\
 &  FS  &  14.29  &  0.0  &  18.18  &  0.0  &  0.0  &  0.0  &  16.67  &  50.0  &  0.0  &  0.0 \\ \hline
\multirow{3}{*}{Mixtral8x7B}  &  ZS  &  0.0  &  0.0  &  50.0  &  33.33  &  66.67  &  0.0  &  0.0  &  0.0  &  50.0  &  0.0 \\ 
 &  CoT  &  0.0  &  0.0  &  0.0  &  50.0  &  50.0  &  0.0  &  0.0  &  25.0  &  0.0  &  0.0 \\ 
 &  FS  &  0.0  &  0.0  &  25.0  &  50.0  &  50.0  &  0.0  &  0.0  &  0.0  &  0.0  &  0.0 \\ \hline
\multirow{3}{*}{Mistral Large}  &  ZS  &  0.0  &  0.0  &  0.0  &  100.0  &  100.0  &  0.0  &  0.0  &  0.0  &  0.0  &  0.0 \\ 
 &  CoT  &  0.0  &  0.0  &  0.0  &  66.67  &  0.0  &  0.0  &  0.0  &  0.0  &  0.0  &  0.0 \\ 
 &  FS  &  0.0  &  0.0  &  0.0  &  100.0  &  50.0  &  0.0  &  0.0  &  0.0  &  0.0  &  0.0 \\ \hline
\multirow{3}{*}{Llama8B}  &  ZS  &  50.0  &  70.0  &  87.5  &  87.5  &  90.0  &  33.33  &  28.57  &  66.67  &  57.14  &  77.78 \\ 
 &  CoT  &  50.0  &  0.0  &  60.0  &  66.67  &  40.0  &  14.29  &  75.0  &  80.0  &  25.0  &  33.33 \\ 
 &  FS  &  0.0  &  0.0  &  0.0  &  0.0  &  0.0  &  0.0  &  0.0  &  0.0  &  0.0  &  0.0 \\ \hline
\multirow{3}{*}{Llama70B}  &  ZS  &  100.0  &  50.0  &  20.0  &  0.0  &  0.0  &  50.0  &  0.0  &  0.0  &  0.0  &  0.0 \\ 
 &  CoT  &  0.0  &  0.0  &  0.0  &  0.0  &  0.0  &  0.0  &  0.0  &  50.0  &  0.0  &  0.0 \\ 
 &  FS  &  0.0  &  0.0  &  0.0  &  0.0  &  0.0  &  0.0  &  0.0  &  0.0  &  0.0  &  0.0 \\ \hline
\multirow{3}{*}{Llama405B}  &  ZS  &  0.0  &  0.0  &  0.0  &  0.0  &  0.0  &  0.0  &  0.0  &  0.0  &  0.0  &  0.0 \\
 &  CoT  &  0.0  &  0.0  &  0.0  &  0.0  &  0.0  &  0.0  &  0.0  &  0.0  &  0.0  &  0.0 \\ 
 &  FS  &  0.0  &  0.0  &  0.0  &  0.0  &  0.0  &  0.0  &  0.0  &  0.0  &  0.0  &  0.0 \\ \hline
 \multirow{3}{*}{Titan lite}  &  ZS & 0.0 & 0.0 & 0.0 & 0.0 & 0.0 & 0.0 & 0.0 & 0.0 & 0.0 & 0.0  \\
 &  CoT & 0.0 & 0.0 & 0.0 & 20.0 & 0.0 & 0.0 & 0.0 & 0.0 & 0.0 & 0.0  \\
 &  FS & 0.0 & 14.29 & 0.0 & 0.0 & 16.67 & 0.0 & 0.0 & 0.0 & 0.0 & 0.0  \\
\hline
\multirow{3}{*}{Titan express}  &  ZS & 0.0 & 0.0 & 0.0 & 0.0 & 0.0 & 0.0 & 0.0 & 0.0 & 0.0 & 0.0  \\
 &  CoT & 0.0 & 0.0 & 0.0 & 28.57 & 0.0 & 0.0 & 0.0 & 0.0 & 0.0 & 0.0  \\
 &  FS & 0.0 & 0.0 & 0.0 & 0.0 & 0.0 & 0.0 & 0.0 & 0.0 & 0.0 & 0.0  \\
\hline
\multirow{3}{*}{Titan large}  &  ZS  &  0.0  &  0.0  &  0.0  &  20.0  &  0.0  &  0.0  &  0.0  &  0.0  &  0.0  &  0.0 \\
 &  CoT  &  0.0  &  0.0  &  0.0  &  0.0  &  0.0  &  0.0  &  0.0  &  0.0  &  0.0  &  0.0 \\ 
 &  FS  &  0.0  &  0.0  &  0.0  &  0.0  &  0.0  &  0.0  &  0.0  &  0.0  &  0.0  &  0.0 \\ \hline
\multirow{3}{*}{Command r}  &  ZS  &  0.0  &  0.0  &  0.0  &  0.0  &  0.0  &  0.0  &  0.0  &  0.0  &  0.0  &  0.0 \\
 &  CoT  &  0.0  &  0.0  &  0.0  &  0.0  &  0.0  &  0.0  &  0.0  &  0.0  &  0.0  &  0.0 \\ 
 &  FS  &  0.0  &  0.0  &  0.0  &  0.0  &  0.0  &  0.0  &  0.0  &  0.0  &  0.0  &  0.0 \\ \hline
\multirow{3}{*}{Command r plus}  &  ZS  &  0.0  &  0.0  &  0.0  &  0.0  &  0.0  &  0.0  &  0.0  &  0.0  &  0.0  &  0.0 \\
 &  CoT  &  0.0  &  0.0  &  0.0  &  0.0  &  0.0  &  0.0  &  0.0  &  0.0  &  0.0  &  0.0 \\ 
 &  FS  &  0.0  &  0.0  &  0.0  &  0.0  &  0.0  &  0.0  &  0.0  &  0.0  &  0.0  &  0.0 \\ \hline
\multirow{3}{*}{Command light text}  &  ZS  &  0.0  &  0.0  &  0.0  &  0.0  &  0.0  &  0.0  &  0.0  &  0.0  &  0.0  &  0.0 \\
 &  CoT  &  0.0  &  0.0  &  0.0  &  0.0  &  0.0  &  0.0  &  0.0  &  0.0  &  0.0  &  0.0 \\ 
 &  FS  &  0.0  &  0.0  &  0.0  &  0.0  &  0.0  &  0.0  &  0.0  &  0.0  &  0.0  &  0.0 \\ \hline
\multirow{3}{*}{Command text}  &  ZS  &  0.0  &  0.0  &  0.0  &  0.0  &  0.0  &  0.0  &  16.67  &  0.0  &  0.0  &  0.0 \\
 &  CoT  &  0.0  &  0.0  &  0.0  &  0.0  &  0.0  &  0.0  &  0.0  &  0.0  &  0.0  &  0.0 \\ 
 &  FS  &  0.0  &  0.0  &  0.0  &  0.0  &  12.5  &  0.0  &  0.0  &  0.0  &  0.0  &  0.0 \\ \hline
\multirow{3}{*}{Claude opus}  &  ZS  &  0.0  &  0.0  &  0.0  &  0.0  &  0.0  &  0.0  &  0.0  &  0.0  &  0.0  &  0.0 \\
 &  CoT  &  0.0  &  0.0  &  0.0  &  0.0  &  0.0  &  0.0  &  0.0  &  0.0  &  0.0  &  0.0 \\ 
 &  FS  &  0.0  &  0.0  &  0.0  &  0.0  &  0.0  &  0.0  &  0.0  &  0.0  &  0.0  &  0.0 \\ \hline
\multirow{3}{*}{Claude instant}  &  ZS  &  0.0  &  0.0  &  0.0  &  0.0  &  0.0  &  0.0  &  0.0  &  0.0  &  0.0  &  0.0 \\
 &  CoT  &  0.0  &  0.0  &  0.0  &  0.0  &  0.0  &  0.0  &  0.0  &  0.0  &  0.0  &  0.0 \\
 &  FS  &  0.0  &  50.0  &  0.0  &  0.0  &  0.0  &  0.0  &  0.0  &  0.0  &  0.0  &  0.0 \\ \hline
\multirow{3}{*}{Claude haiku}  &  ZS  &  0.0  &  0.0  &  0.0  &  0.0  &  0.0  &  0.0  &  0.0  &  0.0  &  0.0  &  0.0 \\
 &  CoT  &  0.0  &  0.0  &  0.0  &  0.0  &  0.0  &  0.0  &  0.0  &  0.0  &  0.0  &  0.0 \\ 
 &  FS  &  0.0  &  0.0  &  0.0  &  0.0  &  0.0  &  0.0  &  0.0  &  0.0  &  0.0  &  0.0 \\ \hline
 \multirow{3}{*}{Claude v2}  &  ZS  &  0.0  &  0.0  &  25.0  &  100.0  &  16.67  &  0.0  &  0.0  &  0.0  &  0.0  &  0.0 \\
 &  CoT  &  0.0  &  0.0  &  14.29  &  100.0  &  0.0  &  0.0  &  100.0  &  0.0  &  0.0  &  50.0 \\
 &  FS  &  0.0  &  0.0  &  0.0  &  100.0  &  0.0  &  0.0  &  0.0  &  0.0  &  0.0  &  33.33 \\ \hline
\multirow{3}{*}{Claude 3.5 Sonnet}  &  ZS  &  0.0  &  0.0  &  0.0  &  0.0  &  0.0  &  0.0  &  0.0  &  0.0  &  0.0  &  0.0 \\
 &  CoT  &  0.0  &  0.0  &  0.0  &  0.0  &  0.0  &  0.0  &  0.0  &  0.0  &  0.0  &  0.0 \\
 &  FS  &  0.0  &  0.0  &  0.0  &  0.0  &  0.0  &  0.0  &  0.0  &  0.0  &  0.0  &  0.0 \\ \hline
\multirow{3}{*}{Claude 3.7 Sonnet}  &  ZS  &  0.0  &  0.0  &  0.0  &  0.0  &  0.0  &  0.0  &  0.0  &  0.0  &  0.0  &  0.0 \\
 &  CoT  &  0.0  &  0.0  &  0.0  &  0.0  &  0.0  &  0.0  &  0.0  &  0.0  &  0.0  &  0.0 \\
 &  FS  &  0.0  &  0.0  &  0.0  &  0.0  &  0.0  &  0.0  &  0.0  &  0.0  &  0.0  &  0.0 \\
 \hline
\end{tabular}
\caption{The refusal rate for each LLM, prompting technique, and type of constants redefinitions for $Q_1$ questions.}
\label{tab:refusal-constants-q1}
\end{table*}

\begin{table*}[h!]
\small
\centering
\begin{tabular}{l|c|ccccc|ccccc}
\hline
\multirow{2}{*}{Model} & \multirow{2}{*}{Prompt} & \multicolumn{5}{c|}{FF}               & \multicolumn{5}{c}{MC}                \\
\cline{3-12}
                       &                         & $R_a1$ & $R_a2$ & $R_a3$ & $R_s1$ & $R_s2$ & $R_a1$ & $R_a2$ & $R_a3$ & $R_s1$ & $R_s2$ \\ \hline

\multirow{3}{*}{Mistral7B}  &  ZS  &  0.0  &  0.0  &  0.0  &  20.0  &  8.33  &  0.0  &  0.0  &  20.0  &  9.09  &  14.29 \\
 &  CoT  &  0.0  &  0.0  &  0.0  &  14.29  &  25.0  &  0.0  &  0.0  &  25.0  &  0.0  &  25.0 \\
 &  FS  &  0.0  &  0.0  &  0.0  &  0.0  &  0.0  &  10.0  &  33.33  &  0.0  &  30.0  &  11.11 \\
\hline
\multirow{3}{*}{Mixtral8x7B}  &  ZS  &  0.0  &  0.0  &  0.0  &  50.0  &  40.0  &  0.0  &  0.0  &  0.0  &  0.0  &  0.0 \\
 &  CoT  &  0.0  &  0.0  &  0.0  &  0.0  &  16.67  &  0.0  &  0.0  &  0.0  &  33.33  &  37.5 \\
 &  FS  &  0.0  &  0.0  &  0.0  &  0.0  &  0.0  &  0.0  &  0.0  &  0.0  &  0.0  &  0.0 \\
\hline
\multirow{3}{*}{Mistral Large}  &  ZS  &  0.0  &  0.0  &  0.0  &  0.0  &  20.0  &  0.0  &  0.0  &  0.0  &  0.0  &  0.0 \\
 &  CoT  &  0.0  &  0.0  &  0.0  &  0.0  &  0.0  &  0.0  &  0.0  &  0.0  &  0.0  &  0.0 \\
 &  FS  &  0.0  &  0.0  &  0.0  &  0.0  &  28.57  &  0.0  &  0.0  &  0.0  &  0.0  &  0.0 \\
\hline
\multirow{3}{*}{Llama8B}  &  ZS  &  50.0  &  55.56  &  88.89  &  40.0  &  50.0  &  28.57  &  20.0  &  50.0  &  11.11  &  33.33 \\
 &  CoT  &  60.0  &  66.67  &  28.57  &  42.86  &  42.86  &  20.0  &  14.29  &  60.0  &  0.0  &  33.33 \\
 &  FS  &  0.0  &  0.0  &  0.0  &  11.11  &  25.0  &  0.0  &  0.0  &  0.0  &  0.0  &  0.0 \\
\hline
\multirow{3}{*}{Llama70B}  &  ZS  &  50.0  &  40.0  &  80.0  &  25.0  &  14.29  &  0.0  &  0.0  &  0.0  &  0.0  &  0.0 \\
 &  CoT  &  50.0  &  0.0  &  0.0  &  50.0  &  0.0  &  0.0  &  0.0  &  100.0  &  0.0  &  0.0 \\
 &  FS  &  0.0  &  0.0  &  0.0  &  0.0  &  0.0  &  0.0  &  0.0  &  0.0  &  0.0  &  0.0 \\
\hline
\multirow{3}{*}{Llama405B}  &  ZS  &  0.0  &  0.0  &  0.0  &  0.0  &  0.0  &  0.0  &  0.0  &  0.0  &  0.0  &  0.0 \\
 &  CoT  &  0.0  &  0.0  &  0.0  &  0.0  &  0.0  &  0.0  &  0.0  &  0.0  &  0.0  &  0.0 \\
 &  FS  &  0.0  &  0.0  &  0.0  &  0.0  &  0.0  &  0.0  &  0.0  &  0.0  &  0.0  &  0.0 \\
\hline
\multirow{3}{*}{Titan lite}  &  ZS & 0.0 & 0.0 & 0.0 & 10.0 & 0.0 & 0.0 & 0.0 & 0.0 & 0.0 & 0.0 \\
 &  CoT & 0.0 & 0.0 & 0.0 & 0.0 & 10.0 & 0.0 & 0.0 & 0.0 & 0.0 & 0.0  \\
 &  FS & 0.0 & 7.14 & 0.0 & 0.0 & 0.0 & 0.0 & 0.0 & 0.0 & 0.0 & 0.0  \\
\hline
\multirow{3}{*}{Titan express}  &  ZS & 0.0 & 0.0 & 0.0 & 0.0 & 0.0 & 0.0 & 0.0 & 0.0 & 0.0 & 0.0  \\
 &  CoT & 0.0 & 0.0 & 0.0 & 0.0 & 0.0 & 0.0 & 0.0 & 0.0 & 0.0 & 0.0  \\
 &  FS & 0.0 & 0.0 & 0.0 & 0.0 & 0.0 & 0.0 & 0.0 & 0.0 & 0.0 & 0.0  \\
\hline

\multirow{3}{*}{Titan large}  &  ZS  &  0.0  &  0.0  &  0.0  &  10.0  &  0.0  &  0.0  &  0.0  &  0.0  &  0.0  &  0.0 \\
 &  CoT  &  0.0  &  0.0  &  0.0  &  0.0  &  0.0  &  0.0  &  0.0  &  0.0  &  0.0  &  0.0 \\
 &  FS  &  0.0  &  0.0  &  0.0  &  0.0  &  0.0  &  0.0  &  0.0  &  0.0  &  0.0  &  0.0 \\
\hline
\multirow{3}{*}{Command r}  &  ZS  &  0.0  &  0.0  &  0.0  &  0.0  &  0.0  &  0.0  &  0.0  &  0.0  &  0.0  &  0.0 \\
 &  CoT  &  0.0  &  0.0  &  0.0  &  0.0  &  0.0  &  0.0  &  0.0  &  0.0  &  0.0  &  0.0 \\
 &  FS  &  0.0  &  0.0  &  0.0  &  0.0  &  0.0  &  0.0  &  0.0  &  0.0  &  0.0  &  0.0 \\
\hline
\multirow{3}{*}{Command r+}  &  ZS  &  0.0  &  0.0  &  0.0  &  0.0  &  0.0  &  0.0  &  0.0  &  0.0  &  0.0  &  0.0 \\
 &  CoT  &  0.0  &  0.0  &  0.0  &  0.0  &  0.0  &  0.0  &  0.0  &  0.0  &  0.0  &  0.0 \\
 &  FS  &  0.0  &  0.0  &  0.0  &  0.0  &  0.0  &  0.0  &  0.0  &  0.0  &  0.0  &  0.0 \\
\hline
\multirow{3}{*}{Command light text}  &  ZS  &  0.0  &  0.0  &  0.0  &  0.0  &  0.0  &  0.0  &  0.0  &  0.0  &  0.0  &  0.0 \\
 &  CoT  &  0.0  &  0.0  &  0.0  &  0.0  &  0.0  &  0.0  &  0.0  &  0.0  &  0.0  &  0.0 \\
 &  FS  &  0.0  &  0.0  &  0.0  &  0.0  &  0.0  &  0.0  &  0.0  &  0.0  &  0.0  &  0.0 \\
\hline
\multirow{3}{*}{Command text}  &  ZS  &  0.0  &  0.0  &  0.0  &  0.0  &  0.0  &  0.0  &  0.0  &  0.0  &  0.0  &  0.0 \\
 &  CoT  &  0.0  &  0.0  &  0.0  &  0.0  &  0.0  &  0.0  &  0.0  &  0.0  &  0.0  &  0.0 \\
 &  FS  &  0.0  &  0.0  &  0.0  &  0.0  &  0.0  &  0.0  &  0.0  &  0.0  &  0.0  &  0.0 \\
\hline
\multirow{3}{*}{Claude opus}  &  ZS  &  0.0  &  0.0  &  0.0  &  0.0  &  0.0  &  0.0  &  0.0  &  0.0  &  0.0  &  0.0 \\
 &  CoT  &  0.0  &  0.0  &  0.0  &  0.0  &  0.0  &  0.0  &  0.0  &  0.0  &  0.0  &  0.0 \\
 &  FS  &  0.0  &  0.0  &  0.0  &  0.0  &  0.0  &  0.0  &  0.0  &  0.0  &  0.0  &  0.0 \\
\hline
\multirow{3}{*}{Claude instant}  &  ZS  &  0.0  &  0.0  &  0.0  &  11.11  &  0.0  &  0.0  &  0.0  &  0.0  &  0.0  &  0.0 \\
 &  CoT  &  0.0  &  0.0  &  0.0  &  0.0  &  0.0  &  0.0  &  0.0  &  0.0  &  0.0  &  0.0 \\
 &  FS  &  0.0  &  0.0  &  0.0  &  11.11  &  0.0  &  0.0  &  0.0  &  0.0  &  0.0  &  0.0 \\
\hline
\multirow{3}{*}{Claude haiku}  &  ZS  &  0.0  &  0.0  &  0.0  &  0.0  &  0.0  &  0.0  &  0.0  &  0.0  &  0.0  &  0.0 \\
 &  CoT  &  0.0  &  0.0  &  0.0  &  0.0  &  0.0  &  0.0  &  0.0  &  0.0  &  0.0  &  0.0 \\
 &  FS  &  0.0  &  0.0  &  0.0  &  0.0  &  0.0  &  0.0  &  0.0  &  0.0  &  0.0  &  0.0 \\
\hline
\multirow{3}{*}{Claude v2}  &  ZS  &  0.0  &  0.0  &  0.0  &  42.86  &  0.0  &  0.0  &  0.0  &  0.0  &  10.0  &  12.5 \\
 &  CoT  &  0.0  &  0.0  &  0.0  &  12.5  &  14.29  &  0.0  &  0.0  &  0.0  &  0.0  &  0.0 \\
 &  FS  &  0.0  &  0.0  &  0.0  &  0.0  &  0.0  &  0.0  &  0.0  &  0.0  &  0.0  &  0.0 \\ \hline
\multirow{3}{*}{Claude 3.5 Sonnet}  &  ZS  &  0.0  &  0.0  &  0.0  &  0.0  &  0.0  &  0.0  &  0.0  &  0.0  &  0.0  &  0.0 \\
 &  CoT  &  0.0  &  0.0  &  0.0  &  0.0  &  0.0  &  0.0  &  0.0  &  0.0  &  0.0  &  0.0 \\
 &  FS  &  0.0  &  0.0  &  0.0  &  0.0  &  0.0  &  0.0  &  0.0  &  0.0  &  0.0  &  0.0 \\
\hline
\multirow{3}{*}{Claude 3.7 Sonnet}  &  ZS  &  0.0  &  0.0  &  0.0  &  0.0  &  0.0  &  0.0  &  0.0  &  0.0  &  0.0  &  0.0 \\
 &  CoT  &  0.0  &  0.0  &  0.0  &  0.0  &  0.0  &  0.0  &  0.0  &  0.0  &  0.0  &  0.0 \\
 &  FS  &  0.0  &  0.0  &  0.0  &  0.0  &  0.0  &  0.0  &  0.0  &  0.0  &  0.0  &  0.0 \\
 \hline
\end{tabular}
\caption{The refusal rate for each LLM, prompting technique, and type of constants redefinition for $Q_2$ questions.}
\label{tab:refusal-constants-q2}
\end{table*}

\begin{table*}[h!]
\small
\centering
\begin{tabular}{l|c|ccccc|ccccc}
\hline
\multirow{2}{*}{Model} & \multirow{2}{*}{Prompt} & \multicolumn{5}{c|}{FF}               & \multicolumn{5}{c}{MC}                \\
\cline{3-12}
                       &                         & $R_a1$ & $R_a2$ & $R_a3$ & $R_s1$ & $R_s2$ & $R_a1$ & $R_a2$ & $R_a3$ & $R_s1$ & $R_s2$ \\ \hline

\multirow{3}{*}{Mistral7B}  &  ZS  &  0.0  &  0.0  &  9.09  &  0.0  &  0.0  &  0.0  &  0.0  &  0.0  &  0.0  &  16.67 \\
 &  CoT  &  0.0  &  0.0  &  0.0  &  0.0  &  9.09  &  40.0  &  14.29  &  14.29  &  16.67  &  9.09 \\
 &  FS  &  0.0  &  0.0  &  23.08  &  0.0  &  0.0  &  0.0  &  0.0  &  0.0  &  0.0  &  0.0 \\
\hline
\multirow{3}{*}{Mixtral8x7B}  &  ZS  &  0.0  &  0.0  &  20.0  &  10.0  &  0.0  &  0.0  &  16.67  &  12.5  &  50.0  &  0.0 \\
 &  CoT  &  0.0  &  0.0  &  0.0  &  12.5  &  9.09  &  0.0  &  33.33  &  20.0  &  33.33  &  50.0 \\
 &  FS  &  0.0  &  14.29  &  11.11  &  14.29  &  0.0  &  14.29  &  0.0  &  0.0  &  0.0  &  75.0 \\
\hline
\multirow{3}{*}{Mistral Large}  &  ZS  &  0.0  &  0.0  &  25.0  &  0.0  &  0.0  &  0.0  &  0.0  &  0.0  &  25.0  &  0.0 \\
 &  CoT  &  0.0  &  0.0  &  25.0  &  0.0  &  33.33  &  0.0  &  0.0  &  0.0  &  0.0  &  0.0 \\
 &  FS  &  0.0  &  0.0  &  20.0  &  0.0  &  16.67  &  0.0  &  0.0  &  0.0  &  0.0  &  20.0 \\
\hline
\multirow{3}{*}{Llama8B}  &  ZS  &  9.09  &  45.45  &  54.55  &  18.18  &  36.36  &  12.5  &  44.44  &  55.56  &  36.36  &  45.45 \\
 &  CoT  &  0.0  &  20.0  &  40.0  &  0.0  &  11.11  &  16.67  &  42.86  &  50.0  &  14.29  &  14.29 \\
 &  FS  &  0.0  &  0.0  &  0.0  &  0.0  &  0.0  &  0.0  &  0.0  &  0.0  &  0.0  &  0.0 \\
\hline
\multirow{3}{*}{Llama70B}  &  ZS  &  50.0  &  75.0  &  55.56  &  0.0  &  20.0  &  33.33  &  0.0  &  0.0  &  0.0  &  0.0 \\
 &  CoT  &  0.0  &  0.0  &  12.5  &  25.0  &  0.0  &  33.33  &  16.67  &  0.0  &  0.0  &  0.0 \\
 &  FS  &  0.0  &  0.0  &  0.0  &  0.0  &  0.0  &  0.0  &  0.0  &  0.0  &  0.0  &  0.0 \\
\hline
\multirow{3}{*}{Llama405B}  &  ZS  &  0.0  &  0.0  &  0.0  &  0.0  &  20.0  &  0.0  &  0.0  &  0.0  &  0.0  &  0.0 \\
 &  CoT  &  0.0  &  0.0  &  0.0  &  0.0  &  0.0  &  0.0  &  0.0  &  0.0  &  0.0  &  0.0 \\
 &  FS  &  0.0  &  0.0  &  0.0  &  0.0  &  0.0  &  0.0  &  0.0  &  0.0  &  0.0  &  0.0 \\
\hline

\multirow{3}{*}{Titan lite}  &  ZS & 0.0 & 0.0 & 6.67 & 6.67 & 0.0 & 0.0 & 0.0 & 0.0 & 0.0 & 0.0  \\
 &  CoT & 0.0 & 7.69 & 7.69 & 0.0 & 0.0 & 0.0 & 0.0 & 0.0 & 0.0 & 0.0  \\
 &  FS & 0.0 & 0.0 & 0.0 & 0.0 & 0.0 & 0.0 & 0.0 & 0.0 & 0.0 & 0.0 \\
\hline
\multirow{3}{*}{Titan express}  &  ZS & 0.0 & 0.0 & 0.0 & 8.33 & 0.0 & 0.0 & 0.0 & 0.0 & 0.0 & 0.0  \\
 &  CoT & 0.0 & 0.0 & 0.0 & 0.0 & 0.0 & 0.0 & 0.0 & 0.0 & 0.0 & 0.0  \\
 &  FS & 0.0 & 0.0 & 0.0 & 0.0 & 0.0 & 0.0 & 0.0 & 0.0 & 0.0 & 0.0  \\
\hline

\multirow{3}{*}{Titan large}  &  ZS  &  0.0  &  0.0  &  0.0  &  0.0  &  0.0  &  0.0  &  0.0  &  0.0  &  0.0  &  0.0 \\
 &  CoT  &  0.0  &  0.0  &  0.0  &  0.0  &  0.0  &  0.0  &  0.0  &  0.0  &  0.0  &  0.0 \\
 &  FS  &  0.0  &  0.0  &  0.0  &  0.0  &  0.0  &  0.0  &  0.0  &  0.0  &  0.0  &  0.0 \\
\hline
\multirow{3}{*}{Command r}  &  ZS  &  0.0  &  0.0  &  0.0  &  0.0  &  0.0  &  0.0  &  0.0  &  0.0  &  0.0  &  0.0 \\
 &  CoT  &  0.0  &  0.0  &  0.0  &  0.0  &  0.0  &  0.0  &  0.0  &  0.0  &  0.0  &  0.0 \\
 &  FS  &  0.0  &  0.0  &  0.0  &  0.0  &  0.0  &  0.0  &  0.0  &  0.0  &  0.0  &  0.0 \\
\hline
\multirow{3}{*}{Command r+}  &  ZS  &  0.0  &  0.0  &  0.0  &  0.0  &  0.0  &  0.0  &  0.0  &  0.0  &  0.0  &  0.0 \\
 &  CoT  &  0.0  &  0.0  &  0.0  &  0.0  &  0.0  &  0.0  &  0.0  &  0.0  &  0.0  &  0.0 \\
 &  FS  &  0.0  &  0.0  &  0.0  &  0.0  &  0.0  &  0.0  &  0.0  &  0.0  &  0.0  &  0.0 \\
\hline
\multirow{3}{*}{Command light text}  &  ZS  &  0.0  &  0.0  &  0.0  &  0.0  &  0.0  &  0.0  &  0.0  &  0.0  &  0.0  &  0.0 \\
 &  CoT  &  0.0  &  0.0  &  0.0  &  0.0  &  0.0  &  0.0  &  0.0  &  0.0  &  0.0  &  0.0 \\
 &  FS  &  0.0  &  0.0  &  0.0  &  0.0  &  0.0  &  0.0  &  0.0  &  0.0  &  0.0  &  0.0 \\
\hline
\multirow{3}{*}{Command text}  &  ZS  &  0.0  &  0.0  &  0.0  &  0.0  &  0.0  &  0.0  &  0.0  &  0.0  &  0.0  &  33.33 \\
 &  CoT  &  0.0  &  0.0  &  0.0  &  0.0  &  0.0  &  0.0  &  0.0  &  0.0  &  0.0  &  0.0 \\
 &  FS  &  0.0  &  0.0  &  0.0  &  0.0  &  0.0  &  0.0  &  0.0  &  0.0  &  0.0  &  0.0 \\
\hline
\multirow{3}{*}{Claude opus}  &  ZS  &  0.0  &  0.0  &  0.0  &  0.0  &  0.0  &  0.0  &  0.0  &  0.0  &  0.0  &  0.0 \\
 &  CoT  &  0.0  &  0.0  &  0.0  &  0.0  &  0.0  &  0.0  &  0.0  &  0.0  &  0.0  &  0.0 \\
 &  FS  &  0.0  &  0.0  &  0.0  &  0.0  &  0.0  &  0.0  &  0.0  &  0.0  &  0.0  &  0.0 \\
\hline
\multirow{3}{*}{Claude instant}  &  ZS  &  0.0  &  0.0  &  0.0  &  14.29  &  0.0  &  0.0  &  0.0  &  0.0  &  0.0  &  0.0 \\
 &  CoT  &  0.0  &  0.0  &  0.0  &  0.0  &  0.0  &  0.0  &  0.0  &  0.0  &  0.0  &  0.0 \\
 &  FS  &  0.0  &  0.0  &  0.0  &  0.0  &  0.0  &  0.0  &  0.0  &  0.0  &  0.0  &  0.0 \\
\hline
\multirow{3}{*}{Claude haiku}  &  ZS  &  0.0  &  0.0  &  0.0  &  0.0  &  0.0  &  0.0  &  0.0  &  0.0  &  0.0  &  0.0 \\
 &  CoT  &  0.0  &  0.0  &  0.0  &  0.0  &  0.0  &  0.0  &  0.0  &  0.0  &  0.0  &  0.0 \\
 &  FS  &  0.0  &  0.0  &  0.0  &  0.0  &  0.0  &  0.0  &  0.0  &  0.0  &  0.0  &  0.0 \\
\hline
\multirow{3}{*}{Claude v2}  &  ZS  &  14.29  &  16.67  &  16.67  &  50.0  &  25.0  &  0.0  &  0.0  &  16.67  &  33.33  &  0.0 \\
 &  CoT  &  0.0  &  14.29  &  14.29  &  20.0  &  25.0  &  0.0  &  0.0  &  0.0  &  0.0  &  0.0 \\
 &  FS  &  0.0  &  0.0  &  12.5  &  11.11  &  10.0  &  0.0  &  0.0  &  14.29  &  0.0  &  0.0 \\ \hline
\multirow{3}{*}{Claude 3.5 Sonnet}  &  ZS  &  0.0  &  0.0  &  0.0  &  0.0  &  0.0  &  0.0  &  0.0  &  0.0  &  0.0  &  0.0 \\
 &  CoT  &  0.0  &  0.0  &  0.0  &  0.0  &  0.0  &  0.0  &  0.0  &  0.0  &  0.0  &  0.0 \\
 &  FS  &  0.0  &  0.0  &  0.0  &  0.0  &  0.0  &  0.0  &  0.0  &  0.0  &  0.0  &  0.0 \\
\hline
\multirow{3}{*}{Claude 3.7 Sonnet}  &  ZS  &  0.0  &  0.0  &  0.0  &  0.0  &  0.0  &  0.0  &  0.0  &  0.0  &  0.0  &  0.0 \\
 &  CoT  &  0.0  &  0.0  &  0.0  &  0.0  &  0.0  &  0.0  &  0.0  &  0.0  &  0.0  &  0.0 \\
 &  FS  &  0.0  &  0.0  &  0.0  &  0.0  &  0.0  &  0.0  &  0.0  &  0.0  &  0.0  &  0.0 \\
 \hline
\end{tabular}
\caption{The refusal rate for each LLM, prompting technique, and type of constants redefinition for $Q_3$ questions.}
\label{tab:refusal-constants-q3}
\end{table*}

To further investigate  LLM anchoring more accurately, we calculate the rate of anchored responses \textit{only} in cases where the LLM indeed attempts to solve the problem, excluding refusal cases. Table \ref{tab:anchored_rate_refusal} presents this pure refusal rate for models in the ZS prompting setup, focusing on the most challenging redefinitions in \textit{assignment} ($R_a3$) and \textit{swapping} ($R_s2$) cases. We exclude LLMs where no refusals occurred, as their results are identical to those reported in Table \ref{tab:anchored_table}.

\begin{table*}[h!]
\centering
\small
\begin{tabular}{l|p{0.7cm}p{0.7cm}|p{0.7cm}p{0.7cm}|p{0.7cm}p{0.7cm}|p{0.8cm}p{0.7cm}|p{0.7cm}p{0.7cm}|p{0.7cm}p{0.6cm}}
\hline
\multirow{3}{*}{Model} & \multicolumn{6}{c|}{$R_a3$}                                                               & \multicolumn{6}{c}{$R_s2$}                                                                \\ \cline{2-13}
                       & \multicolumn{2}{c}{$Q_1$} & \multicolumn{2}{c}{$Q_2$} & \multicolumn{2}{c|}{$Q_3$} & \multicolumn{2}{c}{$Q_1$} & \multicolumn{2}{c}{$Q_2$} & \multicolumn{2}{c}{$Q_3$} \\ \cline{2-13}
                       & FF           & MC           & FF           & MC           & FF            & MC           & FF           & MC           & FF           & MC           & FF           & MC           \\ \hline
Mistral7B & 33.33 & \textbf{50.0} & \textbf{33.33} & \textbf{28.57} & 28.57 & 40.0 & 45.45 & 53.33 & 14.28 & 35.71 & 26.67 & 23.08 \\ 
Mixtral8x7B & \textbf{38.46}& 33.33 & 26.67 & 26.67 & 23.08 & 35.71 & 36.37 & 46.67 & 46.15 & \textbf{53.33} & 46.67 & \textbf{73.33} \\ 
Mistral Large & 33.33 & 20.0 & 26.67 & 26.67 & \textbf{57.14} & \textbf{66.67} & \textbf{71.43} & \textbf{53.33} & \textbf{50.0} & 40.0 & \textbf{73.33} & 66.67 \\  \hline
Llama8B & 0.0 & \textbf{44.45} & 0.0 & \textbf{36.37} & 22.22 & 50.0 &\textbf{50.0} & 24.99 & \textbf{40.0} & \textbf{50.0 }& 27.27 & 30.0 \\ 
Llama70B & \textbf{7.15} & 13.33 & 0.0 & 0.0 & 20.0 & 40.0 & 33.33 & 46.67 & 14.28 & 46.67 & 35.71 & 73.33 \\ 
Llama405B & 0.0 & 0.0 & 0.0 & 13.33 & \textbf{26.67} & \textbf{53.33} & 26.67 & \textbf{46.67} & 6.67 & 20.0 & \textbf{57.14} & \textbf{93.33} \\  \hline
Command text & 13.33  & 20.0  & 6.67 & 6.67 & 6.67 & 26.67 & 40.0 & 26.67  & 13.33 & 26.67 & 13.33 & 38.46 \\  \hline
Claude v2 & 28.57  & 13.33  & 20.0 & 0.0 & 50.0  & 42.86 & 14.28 & 40.0 & 33.33  & 21.43 & 46.15 & 66.67 \\ 
\hline
\end{tabular}

\caption{The percentage of anchored responses for the models in the ZS setup for the most difficult constants redefinitions in \textit{assignment} ($R_a3$) and \textit{swapping} ($R_s2$). The highest number for each model family is presented in \textbf{bold}. We exclude models where no refusals occurred, as their results are identical to those in Table \ref{tab:anchored_table}.}
\label{tab:anchored_rate_refusal}
\end{table*}

\section{Results on units of measure redefinition}
\label{app:units_redefinition}


An overview of response accuracy is presented in Table \ref{tab:anchored_table_units}, where we consider the hardest redefinitions corresponding to the $R_{a}2$ and $R_{a}3$ \textit{assignment} types, as well as all three question levels, together with FF and MC response formats regarding units of measurement redefinitions. Additionally, Table \ref{tab:anchored_table_NR_units} presents the number of correct responses in the NR case alongside the anchoring rate for FF responses regarding units redefinitions. Anchoring is less prominent for some LLMs in comparison to their anchored responses in the constants redefinition task; for instance, Command r+, light text, text achieve even 0\% anchoring in some cases, even in $Q_3$ questions over the hardest $R_a3$ unit redefinitions. Nevertheless, anchoring still persists in many instances, with large rates concerning models in the Mistral family for the hardest question and redefinition levels. Moreover, Titan models present high anchoring even for $R_a2$ unit redefinitions, even in the easier $Q_1$ level. Surprisingly, anchoring for Titan models reduces as questions and redefinitions become harder, but this does not indicate an improvement in producing correct responses and therefore an advancement in reasoning capability; instead, the anchoring reduction is attributed to the generation of more completely wrong responses, indicating those models' inability of solving the unit of measure redefinition task appropriately.

\begin{table*}[h!]
\centering
\small
\begin{tabular}{l|p{0.7cm}p{0.6cm}|p{0.7cm}p{0.6cm}|p{0.7cm}p{0.6cm}|p{0.7cm}p{0.6cm}|p{0.7cm}p{0.6cm}|p{0.7cm}p{0.6cm}}
\hline
\multirow{3}{*}{Model} & \multicolumn{6}{c|}{$R_a2$}                                                               & \multicolumn{6}{c}{$R_a3$}                                                                \\ \cline{2-13}
                       & \multicolumn{2}{c}{$Q_1$} & \multicolumn{2}{c}{$Q_2$} & \multicolumn{2}{c|}{$Q_3$} & \multicolumn{2}{c}{$Q_1$} & \multicolumn{2}{c}{$Q_2$} & \multicolumn{2}{c}{$Q_3$} \\ \cline{2-13}
                       & FF           & MC           & FF           & MC           & FF            & MC           & FF           & MC           & FF           & MC           & FF           & MC           \\ \hline

Mistral7B & 0.0 & 37.5 & 25.0 & 25.0 & 18.75 & 56.25 & \textbf{62.5} & 25.0 & \textbf{31.25} & \textbf{37.5} & 31.25 & 25.0 \\ 
Mixtral8x7B & \textbf{6.25} & 31.25 & \textbf{31.25} & \textbf{37.5} & 31.25 & 37.5 & 6.25 & \textbf{31.25} & 6.25 & 31.25 & \textbf{31.25}  & \textbf{50.0 } \\ 
Mistral Large & 0.0 & \textbf{37.5} & 6.25 & 37.5 & 12.5 & 56.25 & 0.0 & 25.0 & 12.5 & \textbf{37.5} & 12.5 & 43.75 \\ \hline
Llama8B & 0.0 & \textbf{25.0} & \textbf{6.25} & \textbf{31.25} & 12.5 & 31.25 & \textbf{6.25} & \textbf{31.25}  & \textbf{12.5} & \textbf{50.0} & \textbf{25.0} & 50.0 \\ 
Llama70B & 0.0 & 6.25 & \textbf{6.25} & \textbf{31.25} & \textbf{25.0} & \textbf{56.25} & 0.0 & 18.75 & 0.0 & \textbf{50.0} & 12.5 & \textbf{62.5} \\ 
Llama405B & 0.0 & 0.0 & 0.0 & \textbf{31.25} & 12.5 & 37.5 & 0.0 & 0.0 & 6.25 & 25.0 & \textbf{25.0} & 31.25 \\ \hline
Titan lite & 6.25 & \textbf{25.0} & 12.5 & \textbf{31.25} & 12.5 & \textbf{25.0} & 25.0 & \textbf{31.25} & 25.0 & 12.5 & 0.0 & 18.75 \\
Titan express & 18.75 & \textbf{25.0} & \textbf{25.0} & 18.75 & 12.5 & \textbf{25.0} & \textbf{43.75} & 25.0 & 31.25 & 12.5 & 6.25 & 18.75 \\
Titan large & \textbf{31.25} & 12.5 & 12.5 & \textbf{31.25} & \textbf{18.75} & \textbf{25.0} & 25.0 & 12.5 & \textbf{37.5} & \textbf{31.25} & 6.25 & \textbf{25.0} \\ \hline
Command r & \textbf{12.5 }& 18.75 & \textbf{12.5} & \textbf{31.25}  & 25.0 & 18.75 & 6.25 & 25.0 & \textbf{12.5} & 18.75 & \textbf{12.5} & 31.25 \\ 
Command r+ & 6.25 & \textbf{43.75} & 0.0 & 25.0 & \textbf{37.5} & \textbf{50.0} & \textbf{6.25} & \textbf{31.25}  & 0.0 & \textbf{31.25} & 0.0 & 25.0 \\ 
Command light text & 6.25 & 12.5 & 0.0 & 25.0 & 6.25 & 25.0 & 12.5 & 25.0 & 6.25 & 31.25 & 0.0 & \textbf{50.0} \\ 
Command text & 12.5 & 12.5 & 12.5 & 18.75 & 0.0 & 18.75 & 0.0 & 31.25 & 12.5 & 12.5 & 0.0 & 43.75 \\ \hline
Claude opus & 0.0 & 0.0 & 0.0 & 6.25 & 12.5 & 25.0 & 0.0 & 0.0 & 0.0 & 0.0 & 0.0 & 6.25 \\ 
Claude instant & \textbf{6.25} & \textbf{25.0} & \textbf{12.5} & 25.0 & 0.0 & \textbf{43.75} & 0.0 & \textbf{43.75} & 0.0 & \textbf{37.5} & 6.25 & \textbf{31.25} \\ 
Claude haiku & 0.0 & 18.75 & 0.0 & 12.5 & 6.25 & 31.25 & 0.0 & 6.25 & 0.0 & 6.25 & \textbf{18.75} & \textbf{31.25} \\ 
Claude v2 & \textbf{6.25} & 18.75 & 6.25 & \textbf{31.25} & \textbf{18.75} & 31.25 & \textbf{6.25} & 0.0 & \textbf{6.25} & 25.0 & 6.25 & 12.5 \\
Claude 3.5 Sonnet & 0.0 & 0.0 & 0.0 & 12.5 & 6.25 & 6.25 & 0.0 & 0.0 & 0.0 & 6.25 & 0.0 & 0.0 \\ 
Claude 3.7 Sonnet & 0.0 & 0.0 & 0.0 & 0.0 & 0.0 & 0.0 & 0.0 & 0.0 & 0.0 & 0.0 & 0.0 & 0.0 \\
\hline
\end{tabular}

\caption{The percentage of anchored responses for all LLMs tested under the ZS prompting setup for the most difficult units of measure redefinitions ($R_{a}2$ and $R_{a}3$ levels). The highest rate for each model family is presented in \textbf{bold}.}
\label{tab:anchored_table_units}
\end{table*}

\begin{table*}[h!]
\centering
\small
\begin{tabular}{l|p{0.7cm}p{0.6cm}|p{0.7cm}p{0.6cm}|p{0.7cm}p{0.6cm}|p{0.7cm}p{0.6cm}|p{0.7cm}c|p{0.7cm}p{0.6cm}}
\hline
\multirow{3}{*}{Model} & \multicolumn{6}{c|}{$R_a2$}                                                               & \multicolumn{6}{c}{$R_a3$}                                                                \\ \cline{2-13}
                       & \multicolumn{2}{c}{$Q_1$} & \multicolumn{2}{c}{$Q_2$} & \multicolumn{2}{c|}{$Q_3$} & \multicolumn{2}{c}{$Q_1$} & \multicolumn{2}{c}{$Q_2$} & \multicolumn{2}{c}{$Q_3$} \\ \cline{2-13}
                       & NR           & FF           & NR           & FF           & NR            & FF           & NR           & FF           & NR           & FF           & NR           & FF           \\ \hline

Mistral 7B & 81.25 & 0.0 & 56.25 & 25.0 & 43.75 & 18.75 & 81.25 & 62.5 & 56.25 & 31.25 & 43.75 & 31.25 \\ 
Mixtral8x7B & 87.5 & 6.25 & 81.25 & 31.25 & 62.5 & 31.25 & 87.5 & 6.25 & 81.25 & 6.25 & 62.5 & 31.25 \\ 
Mistral Large & 93.75 & 0.0 & 93.75 & 6.25 & 81.25 & 12.5 & 93.75 & 0.0 & 93.75 & 12.5 & 81.25 & 12.5 \\ \hline
Llama8B & 75.0 & 0.0 & 56.25 & 6.25 & 6.25 & 12.5 & 75.0 & 6.25 & 56.25 & 12.5 & 6.25 & 25.0 \\ 
Llama70B & 100.0 & 0.0 & 81.25 & 6.25 & 56.25 & 25.0 & 100.0 & 0.0 & 81.25 & 0.0 & 56.25 & 12.5 \\ 
Llama405B & 100.0 & 0.0 & 93.75 & 0.0 & 56.25 & 12.5 & 100.0 & 0.0 & 93.75 & 6.25 & 56.25 & 25.0 \\ \hline
Titan lite & 37.5 & 6.25 & 18.75 & 12.5 & 6.25 & 12.5 & 37.5 & 25.0 & 18.75 & 25.0 & 6.25 & 0.0 \\
Titan express & 75.0 & 18.75 & 37.5 & 25.0 & 6.25 & 12.5 & 75.0 & 43.75 & 37.5 & 31.25 & 6.25 & 6.25 \\
Titan large & 68.75 & 31.25 & 68.75 & 12.5 & 25.0 & 18.75 & 68.75 & 25.0 & 68.75 & 37.5 & 25.0 & 6.25 \\ \hline
Command r & 75.0 & 12.5 & 56.25 & 12.5 & 18.75 & 25.0 & 75.0 & 6.25 & 56.25 & 12.5 & 18.75 & 12.5 \\ 
Command r+ & 87.5 & 6.25 & 93.75 & 0.0 & 81.25 & 37.5 & 87.5 & 6.25 & 93.75 & 0.0 & 81.25 & 0.0 \\ 
Command light text & 31.25 & 6.25 & 6.25 & 0.0 & 0.0 & 6.25 & 31.25 & 12.5 & 6.25 & 6.25 & 0.0 & 0.0 \\ 
Command text & 62.5 & 12.5 & 50.0 & 12.5 & 25.0 & 0.0 & 62.5 & 0.0 & 50.0 & 12.5 & 25.0 & 0.0 \\ \hline
Claude opus & 100.0 & 0.0 & 75.0 & 0.0 & 56.25 & 12.5 & 100.0 & 0.0 & 75.0 & 0.0 & 56.25 & 0.0 \\
Claude instant & 75.0 & 6.25 & 81.25 & 12.5 & 43.75 & 0.0 & 75.0 & 0.0 & 81.25 & 0.0 & 43.75 & 6.25 \\ 
Claude haiku & 100.0 & 0.0 & 93.75 & 0.0 & 81.25 & 6.25 & 100.0 & 0.0 & 93.75 & 0.0 & 81.25 & 18.75 \\ 
Claude v2 & 93.75 & 6.25 & 68.75 & 6.25 & 25.0 & 18.75 & 93.75 & 6.25 & 68.75 & 6.25 & 25.0 & 6.25 \\ 
Claude 3.5 Sonnet & 100.0 & 0.0 & 87.5 & 0.0 & 87.5 & 6.25 & 100.0 & 0.0 & 87.5 & 0.0 & 87.5 & 0.0 \\ 
Claude 3.7 Sonnet & 100.0 & 0.0 & 87.5 & 0.0 & 93.75 & 0.0 & 100.0 & 0.0 & 87.5 & 0.0 & 93.75 & 0.0 \\ 
\hline
\end{tabular}

\caption{The percentage of correct responses with no redefinition (NR) and the anchored response rate for units of measure redefinitions regarding free-form (FF) responses using ZS prompting.}
\label{tab:anchored_table_NR_units}
\end{table*}

\begin{figure}[h!]
    \centering
    \subfloat[Response breakdown for Mistral models.]{ 
        \includegraphics[width=0.8\linewidth]{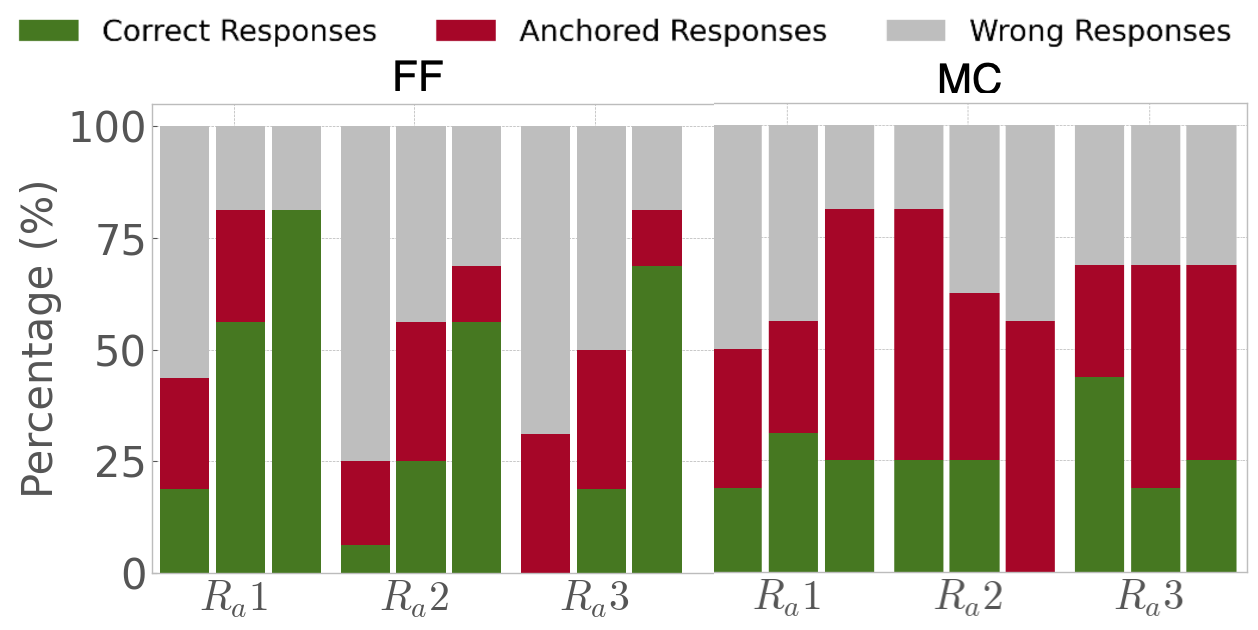}
        \label{fig:mistral}
    }   \\
    \subfloat[Response breakdown for Llama models.]{ 
        \includegraphics[width=0.8\linewidth]{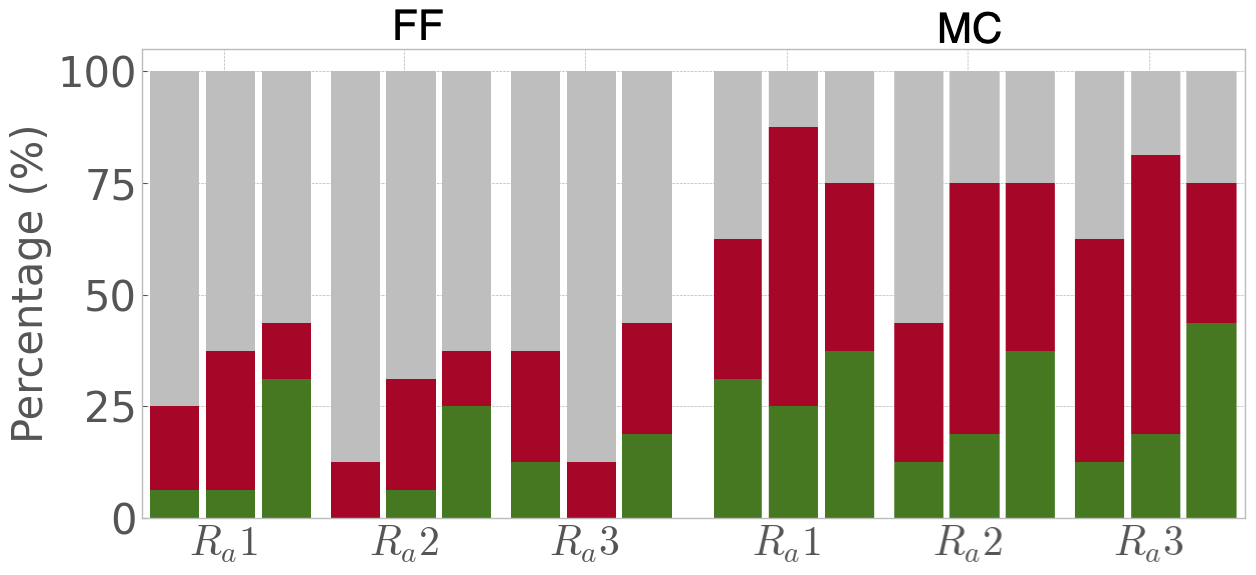}
        \label{fig:llama}
    } \\
        \vskip -0.01in
    \caption{Results for the different Mistral and Llama models on $Q_3$ questions using ZS prompting for the
redefinition task of units of measure redefinitions. The order of the bars per redefinition type/level corresponds to increasing model size.}
    \label{fig:size_comparison_units}
\end{figure}

Figure \ref{fig:size_comparison_units} shows the results of the different Mistral and Llama models for the $Q_3$ question level in the ZS prompting setup for units redefinitions. The conclusions are similar to those for the redefinition of constants, where the number of anchored responses is significantly higher in the MC setup compared to the FF setup --a rather expected pattern, since LLMs are exposed to the default response. Additionally, once again, it is observable that the larger models are more prone to providing anchored responses compared to the smaller ones, regardless the response format and the model family.

\begin{figure*}[h]
\begin{subfigure}{\textwidth}
        \centering
         \includegraphics[width=0.86\linewidth]{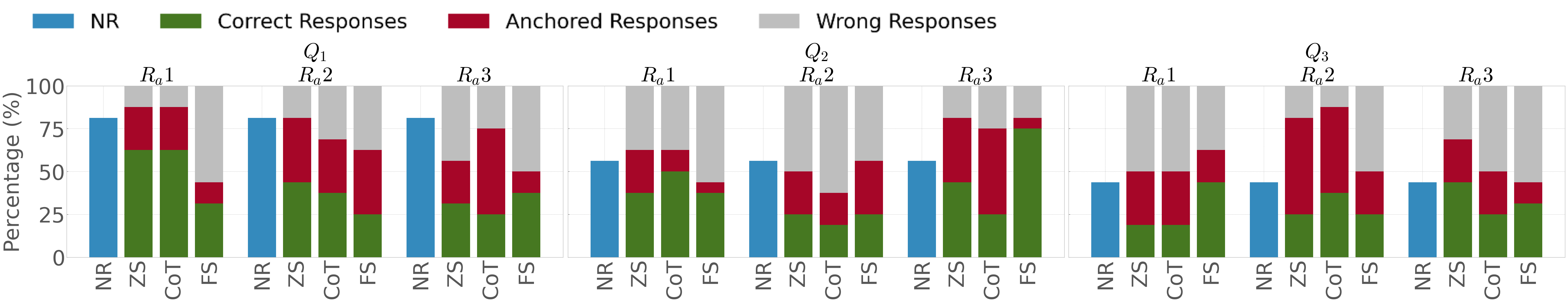}
        \caption{Response breakdown for Mistral 7B before and after units of measure redefinitions.}
        \label{fig:mistral7b_MC-units}
    \end{subfigure}
    
    \begin{subfigure}{\textwidth}
        \centering
        \includegraphics[width=0.86\linewidth]{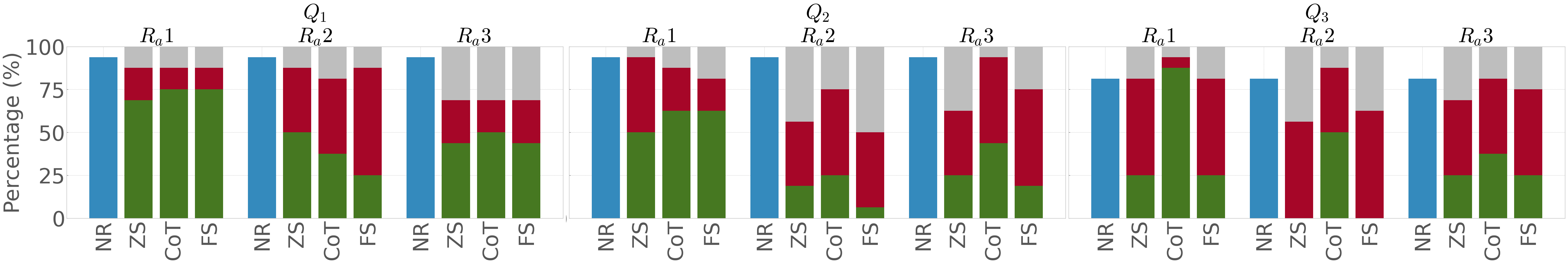}
        \caption{Response breakdown for Mistral Large before and after units of measure redefinitions.}
        \label{fig:mistral_large_MC-units}
    \end{subfigure}
    \caption{Comparison of Mistral 7B and Mistral Large (123B)  responses on the MC response format for units of measure redefinitions.}
    \label{fig:mistral_all-units}
\end{figure*}

\begin{figure*}[h]
\begin{subfigure}{\textwidth}
        \centering
         \includegraphics[width=0.86\linewidth]{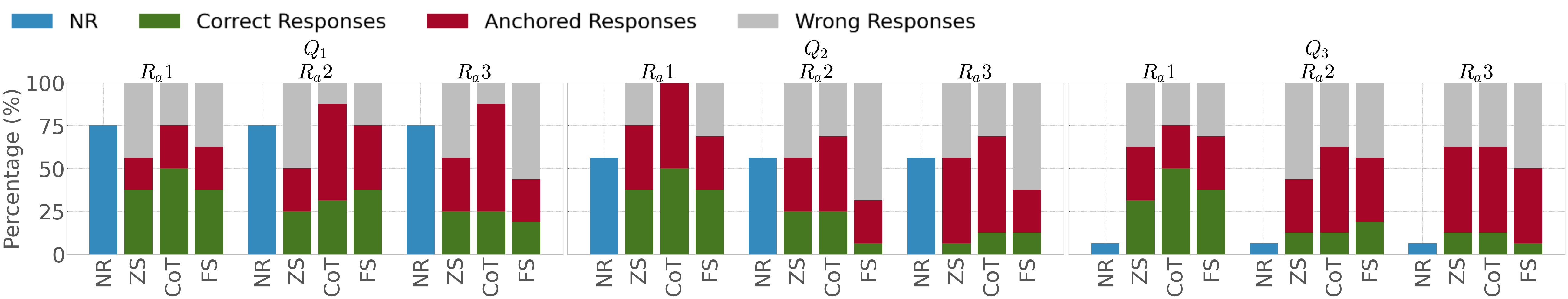}
        \caption{Response breakdown for Llama 8B before and after units of measure redefinitions.}
        \label{fig:llama8b_MC-units}
    \end{subfigure}
    
    \begin{subfigure}{\textwidth}
        \centering
        \includegraphics[width=0.86\linewidth]{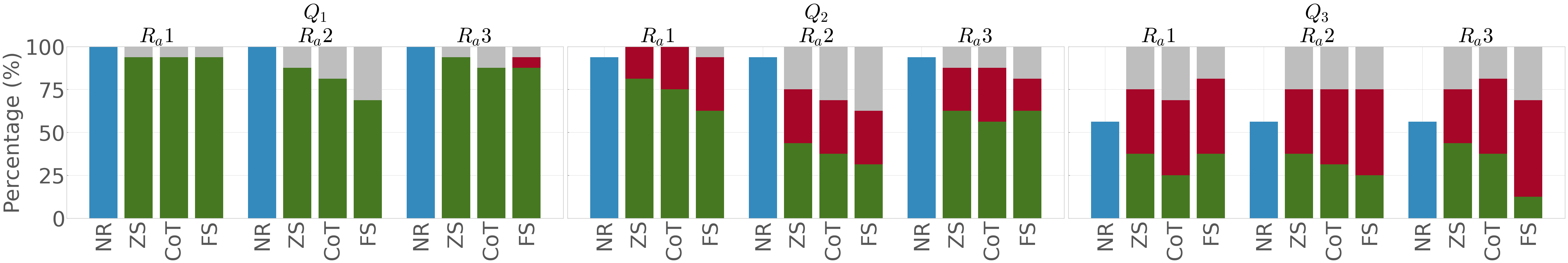}
        \caption{Response breakdown for Llama 405B before and after units of measure redefinitions.}
        \label{fig:llama405b_large_MC-units}
    \end{subfigure}
    \caption{Comparison of Llama8B and Llama405B  responses on the MC response format for units of measure redefinitions.}
    \label{fig:llama_all-units}
\end{figure*}

Furthermore, Figures \ref{fig:mistral_all-units} and \ref{fig:llama_all-units} illustrate the results of Mistral 7B and Mistral Large, as well as Llama 8B and Llama 405B, respectively, regarding units of measure redefinitions. Once again, Mistral 7B tends to provide \textit{fewer anchored responses} compared to its larger counterpart. The same trend holds for Llama, although for $Q_1$ and $Q_2$ responses, the increase in anchoring is less pronounced for the larger model. 

Additionally, the performance of the larger LLama 405B model in the NR case is excellent in the $Q_1$  question level, achieving a response accuracy close to 100\% in most cases, denoting that this model is adequately knowledgeable regarding the default meanings of units of measure. For $Q_2$ questions in Llama 405B, an interesting pattern emerges. The number of correct responses in the NR task remains close to 100\%, indicating that the model is also an excellent reasoner in this difficulty level. However, when units are redefined, the model's accuracy \textit{declines considerably}, coinciding with a noticeable increase in the number of anchored responses. Therefore, Llama 405B exploits memorized patterns to be able to handle unit redefinitions, even though memorization almost useless in the $Q_1$ level. The anchoring rate further increases in the $Q_3$ level, even though the NR correct response rate (and therefore the model's reasoning ability in the default setting) is decreased in comparison to the easier question levels.

Lastly, Tables \ref{tab:correlation_zs-units}, \ref{tab:correlation_fs-units}, and \ref{tab:correlation_cot-units} present the correlation between average model performance for all LLMs in the NR case and the number of anchored responses for unit of measure redefinitions using ZS, FS, and CoT prompting, respectively. These correlations mostly exhibit a similar pattern to those observed in the constant redefinition task. However, unlike constants, where high positive correlations were found due to \textit{swapping}, unit of measure redefinitions are implemented using  \textit{assignment} exclusively. 

Nevertheless, in both ZS and FS prompting, we observe a high positive correlation for $Q_3$-level questions, similar to the constant redefinition case, denoting increased anchoring for more potent reasoners. This trend holds for the MC response format, but not for the FF format (where correlations are weak), contradicting constants redefinition findings. Apparently, anchoring becomes less prominent with respect to reasoning capability when the LLMs have to generate responses over redefined units of measure.

On the other hand, CoT is evidently capable of reducing anchoring of more potent reasoners, leading to weak or negative correlations in all cases, regardless the redefinition or question difficulty. This is  a contradictory fact in comparison to constants redefinitions, revealing that CoT can assist LLMs in reasoning more properly and thus anchor less to their prior knowledge, aligning to basic CoT claims \cite{step-by-step}.

\begin{table}[h!]
\small
    \centering
    \begin{tabular}{l|ccc}
        \hline
        Level & $R_{a1}$ & $R_{a2}$ & $R_{a3}$ \\ \hline
        & \multicolumn{3}{c}{Free-Form (FF)} \\ 
        \hline
        $Q_1$ & -0.295 & \cellcolor{lightdustygreen} -0.403   & \cellcolor{lightdustygreen} -0.33 \\ 
        $Q_2$ & \cellcolor{lightdustygreen} -0.361 & -0.247 & \cellcolor{lightdustygreen} -0.479 \\ 
        $Q_3$ & -0.063 & 0.19 & 0.14 \\
        
        \hline
        & \multicolumn{3}{c}{Multiple Choice (MC)} \\ \hline
        $Q_1$ & \cellcolor{lightdustygreen} -0.49 & -0.149  & \cellcolor{lightdustygreen} -0.542 \\ 
        $Q_2$ & -0.159 & -0.023 & 0.08 \\ 
        $Q_3$ & 0.248 & \cellcolor{lightdustypink} 0.338 & -0.127 \\

        \hline
    \end{tabular}
    \caption{Correlation between model performance before redefinition with the percentage of anchored answers for each type of unit of measure redefinition and question level in ZS setup. 
    Cells highlighted in \textcolor{lightdustypink}{pink} indicate a \textbf{high positive correlation} ($>0.3$), while cells in \textcolor{lightdustygreen}{green} indicate a \textbf{high negative correlation} ($<-0.3$).}
    \label{tab:correlation_zs-units}
\end{table}

\begin{table}[h!]
\small
    \centering
    \begin{tabular}{l|ccc}
        \hline
        Level & $R_{a1}$ & $R_{a2}$ & $R_{a3}$ \\ \hline
        & \multicolumn{3}{c}{Free-Form (FF)} \\ 
        \hline
        $Q_1$ & \cellcolor{lightdustygreen} -0.32 & \cellcolor{lightdustygreen} -0.442 & -0.161 \\
        $Q_2$ & \cellcolor{lightdustygreen} -0.404 & -0.231 & 0.039 \\
        $Q_3$ & 0.128 & -0.042 & 0.279 \\
        
        \hline
                & \multicolumn{3}{c}{Multiple Choice (MC)} \\ \hline

$Q_1$ & \cellcolor{lightdustygreen} -0.332 & 0.058 & \cellcolor{lightdustygreen} -0.593 \\
$Q_2$ & 0.135 & 0.131 & 0.266 \\
$Q_3$ & \cellcolor{lightdustypink} 0.314 & \cellcolor{lightdustypink} 0.49 & 0.101 \\

        \hline
    \end{tabular}
    \caption{Correlation between model performance before redefinition with the percentage of anchored answers for each type of unit of measure redefinition and question level in FS setup. 
    Cells highlighted in \textcolor{lightdustypink}{pink} indicate a \textbf{high positive correlation} ($>0.3$), while cells in \textcolor{lightdustygreen}{green} indicate a \textbf{high negative correlation} ($<-0.3$).}
    \label{tab:correlation_fs-units}
\end{table}

\begin{table}[h!]
\small
    \centering
    \begin{tabular}{l|ccc}
        \hline
        Level & $R_{a1}$ & $R_{a2}$ & $R_{a3}$ \\ \hline
        & \multicolumn{3}{c}{Free-Form (FF)} \\ 
        \hline
$Q_1$ & \cellcolor{lightdustygreen} -0.502 & \cellcolor{lightdustygreen} -0.598 & \cellcolor{lightdustygreen} -0.529 \\
$Q_2$ & \cellcolor{lightdustygreen} -0.465 & \cellcolor{lightdustygreen} -0.3 & -0.174 \\ 
$Q_3$ & -0.232 & -0.181 & -0.079 \\ 
        
        \hline
                & \multicolumn{3}{c}{Multiple Choice (MC)} \\ \hline

$Q_1$ & \cellcolor{lightdustygreen} -0.528 & -0.023 & \cellcolor{lightdustygreen} -0.523 \\ 
$Q_2$ & 0.015 & -0.091 & -0.016 \\
$Q_3$ & -0.127 & 0.013 & -0.242 \\

        \hline
    \end{tabular}
    \caption{Correlation between model performance before redefinition with the percentage of anchored answers for each type of unit of measure redefinition and question level in CoT setup. 
    Cells highlighted in \textcolor{lightdustypink}{pink} indicate a \textbf{high positive correlation} ($>0.3$), while cells in \textcolor{lightdustygreen}{green} indicate a \textbf{high negative correlation} ($<-0.3$).}
    \label{tab:correlation_cot-units}
\end{table}

\newpage
\section{Implementation details}
\label{sec:cards}

We list model cards regarding our employed LLMs in Table \ref{tab:model-cards}. All these LLMs are available in Amazon Bedrock\footnote{\href{https://aws.amazon.com/bedrock/}{https://aws.amazon.com/bedrock/}}, a model deployment service provided by Amazon Web Services (AWS),  accessed via APIs. The code implementing the API calls, as well as the evaluation part of LLM responses is developed in Kaggle notebooks.
\begin{table*}[h!]
\centering \small
\begin{tabular}{p{3.4cm}|p{4.7cm}p{6.7cm}}
\hline
Model name & Model card & URL\\
\hline
Llama8B&  meta-llama/Llama-3.1-8B-Instruct & \href{https://huggingface.co/meta-llama/Llama-3.1-8B-Instruct}{https://huggingface.co/meta-llama/Llama-3.1-8B-Instruct}
\\\hline
Llama70B & meta-llama/Llama-3.1-70B-Instruct & \href{https://huggingface.co/meta-llama/Llama-3.1-70B-Instruct}{https://huggingface.co/meta-llama/Llama-3.1-70B-Instruct} \\\hline
Llama405B & meta-llama/Llama-3.1-405B-Instruct & \href{https://huggingface.co/meta-llama/Llama-3.1-405B-Instruct}{https://huggingface.co/meta-llama/Llama-3.1-405B-Instruct} \\\hline
Mistral7B & mistralai/Mistral-7B-Instruct-v0.2 & \href{https://huggingface.co/mistralai/Mistral-7B-Instruct-v0.2}{https://huggingface.co/mistralai/Mistral-7B-Instruct-v0.2} \\\hline
Mixtral8x7B & mistralai/Mixtral-8x7B-v0.1 & \href{https://huggingface.co/mistralai/Mixtral-8x7B-v0.1}{https://huggingface.co/mistralai/Mixtral-8x7B-v0.1} \\\hline
Mistral Large (123B) & mistral.mistral-large-2402-v1:0& N/A\\\hline
Claude instant v1 & anthropic/claude-instant-1 & \href{https://openrouter.ai/anthropic/claude-instant-1}{https://openrouter.ai/anthropic/claude-instant-1} \\\hline
Claude v2 & anthropic/claude-2 & \href{https://openrouter.ai/anthropic/claude-2}{https://openrouter.ai/anthropic/claude-2}\\\hline
Claude 3 Opus & anthropic/claude-3-opus & \href{https://openrouter.ai/anthropic/claude-3-opus}{https://openrouter.ai/anthropic/claude-3-opus} \\\hline
Claude 3 Haiku & anthropic/claude-3-haiku & \href{https://openrouter.ai/anthropic/claude-3-haiku}{https://openrouter.ai/anthropic/claude-3-haiku}\\\hline
Claude 3.5 Sonnet & anthropic/claude-3.5-sonnet & \href{https://openrouter.ai/anthropic/claude-3.5-sonnet}{https://openrouter.ai/anthropic/claude-3.5-sonnet} \\\hline
Claude 3.7 Sonnet & anthropic/claude-3.7-sonnet & \href{https://openrouter.ai/anthropic/claude-3.7-sonnet}{https://openrouter.ai/anthropic/claude-3.7-sonnet} \\\hline
Cohere command light & cohere.command-light-text-v14 & N/A\\\hline
Cohere command text& cohere.command-text-v14 & N/A\\\hline
Cohere command r & CohereForAI/c4ai-command-r-v01 & \href{https://huggingface.co/CohereForAI/c4ai-command-r-v01}{https://huggingface.co/CohereForAI/c4ai-command-r-v01} 
\\\hline
Cohere command r+ & CohereForAI/c4ai-command-r-plus & \href{https://huggingface.co/CohereForAI/c4ai-command-r-plus}{https://huggingface.co/CohereForAI/c4ai-command-r-plus}
\\\hline
Amazon Titan text lite & amazon.titan-text-lite-v1 & N/A \\\hline
Amazon Titan express & amazon.titan-text-express-v1 & N/A \\\hline
Amazon Titan Tg1 & amazon.titan-tg1-large & N/A \\
\hline
    \end{tabular}
    \caption{Model cards and hyperlinks for used LLMs. N/A stands for not available hyperlink.}
    \label{tab:model-cards}
\end{table*}

Finally, all redefinitions are implemented manually by the authors based on engineering textbooks, with the aid of ChatGPT\footnote{\href{https://chatgpt.com/}{https://chatgpt.com/}} in defining the constants/units to be redefined and as a general guideline towards designing redefinition and question levels. 

\newpage
\section{Prompts}
\label{app:prompts}

This section illustrates the prompts used to question the LLMs. The prompts vary based on the task (NR or redefinition), the required response format (FF or MC), and the prompting techniques selected (ZS, FS, or CoT).

The prompts for the NR task in FF response format are presented below.
 
\par\noindent\rule{0.5\textwidth}{0.4pt}
\\
\small \textbf{prompt\_NR\_FF\_ZS} = `````` 
Answer the following question:\\
<question> \\ 
End the response with the phrase "The final answer is: " followed only by the correct result, with no additional text or commentary.
``````
\\ \\
\small \textbf{prompt\_NR\_FF\_CoT}= ``````Answer the following question:\\
<question> \\ 
Let's think step by step. \\
End the response with the phrase "The final answer is: " followed only by the correct result, with no additional text or commentary.
``````
\\ \\
\small \textbf{prompt\_NR\_FF\_FS} = ``````Answer the following question:\\
<question> \\ 
Here are some examples of similar questions with their correct answers: \\
<NR FF examples> \\
End the response with the phrase "The final answer is: " followed only by the correct result, with
no additional text or commentary.
``````
\par\noindent\rule{0.5\textwidth}{0.4pt}

\normalsize 

Below are the prompts for the NR task regarding the MC response format.

\par\noindent\rule{0.5\textwidth}{0.4pt}
\\
\small \textbf{prompt\_NR\_MC\_ZS} = `````` 
Choose A, B, C or D to answer the question: \\
Question: <question> \\
A: <A> \\
B: <B> \\
C: <C> \\
D: <D> \\
Provide only the letter corresponding to the correct answer: "A", "B", "C", or "D". \\
End the response with the phrase "The final answer is: " followed by the correct letter, with no additional text or commentary.
``````
\\ \\
\small \textbf{prompt\_NR\_MC\_CoT}= ``````Choose A, B, C or D to answer the question: \\
Question: <question> \\
A: <A> \\
B: <B> \\
C: <C> \\
D: <D> \\
Let's think step by step. \\
Provide only the letter corresponding to the correct answer: "A", "B", "C", or "D". \\
End the response with the phrase "The final answer is: " followed by the correct letter, with no additional text or commentary.
``````
\\ \\
\small \textbf{prompt\_NR\_MC\_FS} = ``````Choose A, B, C or D to answer the question: \\
Question: <question> \\
A: <A> \\
B: <B> \\
C: <C> \\
D: <D> \\
Here are some examples of similar questions with their correct answers: \\ 
<NR MC examples> \\
Provide only the letter corresponding to the correct answer: "A", "B", "C", or "D". \\
End the response with the phrase "The final answer is: " followed by the correct letter, with no additional text or commentary.
``````
\par\noindent\rule{0.5\textwidth}{0.4pt}
\normalsize 

For the redefinition tasks, the prompts are identical, with the only difference being the addition of: \textit{``Redefine <X> as <Y>.\textbackslash n''} at the beginning of each prompt. Additionally, in the few-shot setup, the provided examples included redefined values (constants or units, respectively).

The answers generated by the LLMs are parsed through a different model (particularly Claude 3.5 Sonnet) to determine whether they match the correct response or the anchored one. This approach is necessary because LLMs can produce additional outputs (e.g. using CoT), making it difficult to extract their answers using regular expressions. The prompts used in this procedure for the NR and redefinition tasks are presented below.

\par\noindent\rule{0.5\textwidth}{0.4pt}
\small \textbf{prompt\_evaluation\_NR} = ``````You are tasked with comparing two answers: one provided by an LLM (the "LLM answer") and the correct answer (the "real answer"). Your job is to determine if the LLM answer matches the real
answer. \\
The comparison should strictly focus on whether the LLM final answer conveys the same meaning or provide the same information as the correct answer. Minor diﬀerences in phrasing, wording, or structure are acceptable as long as the core meaning remains identical. For numerical results, diﬀerences due to rounding are acceptable as long as the values are reasonably close and within an acceptable margin of error.\\
***Instructions:** \\
1. Compare the LLM answer to the real answer carefully.\\
2. If the LLM answer matches the real answer, output:
correct \\
3. If the LLM answer does not match the real answer, output: incorrect \\
Do not output explanations, justifications, or any additional text — only the word "correct" or "incorrect". \\
***LLM Answer:** \\
<llm\_answer> \\
***Real Answer:** \\
<real\_answer> \\
***Output:** \\
``````

\small \textbf{prompt\_evaluation\_Redefinition} = ``````You are tasked with comparing an answer provided by an LLM (the "LLM answer") to two reference answers: "Reference Answer 1" and "Reference Answer 2". Your job is to determine if the LLM answer matches either of the two reference answers. \\
The comparison should strictly focus on whether the LLM final answer conveys the same meaning or provides the same information as one of the reference answers. Minor diﬀerences in phrasing, wording, or structure are acceptable as long as the core meaning remains identical. For numerical results, diﬀerences due to rounding are acceptable as long as the values are reasonably close and within an acceptable margin of error. \\
***Instructions:** \\
1. Compare the LLM answer carefully with "Reference Answer 1" and "Reference Answer 2". \\
2. If the LLM answer matches "Reference Answer 1", output:
first \\
3. If the LLM answer matches "Reference Answer 2", output:
second \\
4. If the LLM answer matches neither of the two, output:
none \\
Do not output explanations, justifications, or any additional text — only the words "first", "second", or "none". \\
***LLM Answer:** \\
<llm\_answer> \\ 
***Reference Answer 1:** \\
<reference\_answer\_1> \\ 
***Reference Answer 2:** \\ 
<reference\_answer\_2> \\
***Output:** \\
``````
\par\noindent\rule{0.5\textwidth}{0.4pt}
\normalsize

Lastly, to categorize the answers as completely wrong, refused, or blank, we prompt Claude 3.5 Sonnet using the following prompt.

\par\noindent\rule{0.5\textwidth}{0.4pt}
\small

\small \textbf{prompt\_evaluation\_refusal} = ``````You are tasked with analyzing an LLM answer that does not match either of two reference answers: "Reference Answer 1" and "Reference Answer 2". Your job is to classify the LLM answer into one of the following categories: \\
1. ***Wrong Answer**: The LLM provided an incorrect response to the question, either factually or
logically. \\
2. ***Blank Answer**: The LLM provided no substantive response, leaving the answer blank or
completely empty. \\
3. ***Refusal to Answer**: The LLM explicitly refused to answer the question, citing reasons such as the question being nonsensical, impossible to answer, or against its guidelines. \\
***Instructions:** \\
1. Analyze the LLM answer and determine which of the three categories it belongs to. \\
2. If the LLM answer is a **Wrong Answer**, output: wrong \\
3. If the LLM answer is a **Blank Answer**, output: blank \\
4. If the LLM answer is a **Refusal to Answer**, output: refusal \\
5. If the classification is unclear, choose the category that best fits the content of the LLM answer. \\
Do not output explanations, justifications, or any additional text — only the words "wrong", "blank", or "refusal". \\
***LLM Answer:** \\
<llm\_answer> \\
***Reference Answer 1:** \\ 
<reference\_answer\_1> \\
***Reference Answer 2:** \\
<reference\_answer\_2> \\
***Output:**
``````

\par\noindent\rule{0.5\textwidth}{0.4pt}
\normalsize

\end{document}